\documentclass{article}
\usepackage{amssymb}
\usepackage{amsmath}
\usepackage{graphicx}
\usepackage{booktabs}
\usepackage{threeparttable}
\usepackage{wrapfig}
\usepackage{algorithm}
\usepackage{enumitem} 
\usepackage{algorithmic}
\usepackage{appendix}
\usepackage{multirow}
\usepackage{bbm}
\usepackage{siunitx}
\usepackage[table]{xcolor}
\usepackage{subcaption}
\PassOptionsToPackage{numbers,sort&compress}{natbib}
\usepackage[table]{xcolor}
\definecolor{cvprbest}{RGB}{217,232,222}    
\definecolor{cvprsecond}{RGB}{229,235,246}  
\definecolor{cvprheader}{RGB}{248,249,250}  
\usepackage{pifont}

\usepackage[final]{corl_2026} 

\title{Instant-Fold: In-Context Imitation Learning for Deformable Object Manipulation}

\author{
Yilong Wang \\
The Robot Learning Lab \\
Imperial College London \\
yw14218@ic.ac.uk
\And
Cheng Qian \\
The Robot Learning Lab \\
Imperial College London \\
c.qian24@imperial.ac.uk
\And
Edward Johns \\
The Robot Learning Lab \\
Imperial College London \\
e.johns@imperial.ac.uk
}
\begin{document}
\maketitle


\begin{abstract}
Deformable object manipulation (DOM) is challenging due to high-dimensional, partially observable states that evolve through long-horizon, topology-changing interactions with multiple valid manipulation modes. We introduce \emph{Instant-Fold}, an in-context imitation learning framework for DOM. Given a single human demonstration, our policy infers and executes diverse manipulation modes directly from the demonstration—including variations in spatial execution and ordering—without requiring gradient updates. Our approach first learns deformation-aware visual representations via temporal contrastive pretraining, after which a flow-matching transformer policy conditioned on the demonstration predicts actions to execute the intended manipulation mode. Trained entirely in simulation, \emph{Instant-Fold} generalizes across diverse folding modes and transfers zero-shot to real-world settings without additional data collection or fine-tuning.
Videos are available at \url{https://instant-fold.github.io}.
\end{abstract}

\keywords{Deformable Object Manipulation, In-Context Imitation Learning, Contrastive Representation Learning} 

\begin{figure}[h!]
    \centering
    \includegraphics[width=.995\linewidth]{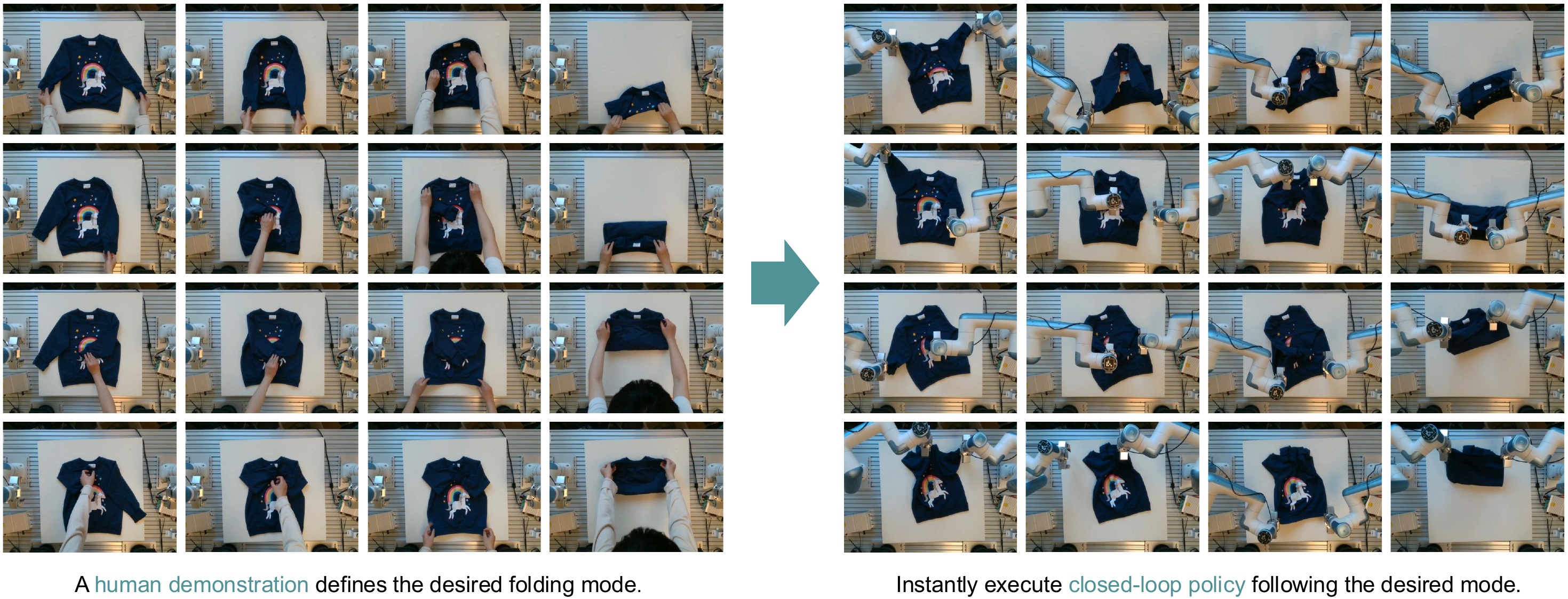}
    \caption{Given a single human demonstration as a prompt, \textsc{Instant-Fold} infers and executes diverse manipulation modes directly from the demonstration without requiring gradient updates.}

    \label{fig:teaser}
\end{figure}

\section{Introduction}

Deformable objects have high-dimensional, partially observable states that evolve through long-horizon, topologically complex interactions. As a result, they admit multiple valid manipulation modes depending on personal preferences, as shown in Figure~\ref{fig:teaser}. The differences are reflected not only in the final configuration, but also in the sequence of intermediate deformations required to achieve it. While such procedures can be specified in language, they are difficult to specify precisely and compactly. In practice, humans rarely rely on verbal instructions alone when teaching complex deformable skills; instead, demonstrations, often in the form of videos, are used to communicate temporal structure, intermediate goals, and subtle spatial conventions that are cumbersome to verbalize. This observation motivates our central premise: for DOM, a single demonstration can provide a richer task interface than language, particularly when the goal is to specify \emph{how} an object should be manipulated rather than merely \emph{what} final state should be achieved. By training on a sufficiently diverse set of modes, a model can, in principle, \emph{learn to infer and reproduce the same mode from a single unseen demonstration} via the emergence of In-Context Learning (ICL) \citep{brown2020language}.

A key challenge in making this formulation work is representation. Foundational vision models are typically trained on large-scale internet images under implicit i.i.d. assumptions \citep{simeoni2025dinov3}, which do not naturally extend to DOM. As an object deforms, its visual appearance and feature representation can change dramatically even when the underlying physical entity remains the same. While such variability may be tolerable when learning a single fixed mode, it becomes a major obstacle for generalization: different valid manipulation modes can induce substantially different deformation trajectories, which a model may incorrectly interpret as entirely different tasks rather than variations within the same task family. A further challenge is data. ICL in language and vision emerged from massive, diverse datasets, yet no comparable corpus exists for DOM. Generalizing to unseen deformation trajectories requires exposure to broad variation in object geometry, intermediate states, and manipulation modes, but the space of possible deformations is effectively unbounded, making such coverage prohibitively expensive and difficult to specify in the real world.

Motivated by these challenges, we propose \textsc{Instant-Fold}, an In-Context Imitation Learning (ICIL) framework for DOM. Our approach explicitly separates \emph{topological} and \emph{functional} reasoning: we first learn deformation-aware representations via temporal contrastive pretraining, capturing correspondences that persist under non-rigid transformations; we then learn reactive control with a flow-matching transformer that conditions on demonstrations to infer task intent. Both components are trained entirely in simulation. We instantiate our framework on clothes folding tasks, where a single object category admits multiple valid modes with distinct temporal structures.

Our contributions are threefold: (1) we formulate DOM as an ICIL problem and introduce a flow-matching transformer architecture for long-horizon deformable in-context manipulation; (2) we propose a temporal contrastive pretraining method tailored to ICIL for learning topology-consistent representations; and (3) we demonstrate successful sim-to-real transfer without any real-world data.
\section{Related Work}
\label{sec:citations}

\paragraph{Representation for Deformable Objects.}
Representations for deformable objects can be broadly grouped into particle-based~\citep{zhang2024adaptigraph,chendaxbench,lin2021softgym,bender2015position}, mesh-based~\citep{pfaff2020learning,10.1145/2994258.2994272}, keypoint-based~\citep{canberk2023cloth,ma2022learning,deng2025general,Wu_2024_CVPR,lips2024learning}, and learned latent or implicit state representations~\citep{lippi2020latent,yan2021learning,zhang2024particle,song2025implicit, tian2025diffusion}. Keypoint-based representations are widely adopted for downstream planning and control \citep{canberk2023cloth, ma2022learning, Wu_2024_CVPR, deng2025general} due to their compactness and computational efficiency. However, existing keypoints are typically either geometric~\citep{shi2021skeleton,zhou2023clothesnet} or semantic~\citep{canberk2023cloth,deng2025general}, limiting their ability to simultaneously preserve spatial fidelity and task-relevant structure, while their sparsity can lead to loss of fine-grained details and sensitivity to occlusions. In contrast, we represent deformable objects as a compact set of geo-semantic tokens, each consisting of a 3D position and a pretrained deformation-aware semantic feature, processed within a transformer architecture.

\paragraph{Sim-to-Real Clothes Folding.}
A major line of work addresses sim-to-real clothes folding using parameterized action primitives such as pick, place, and fling, where policies predict grasp points or affordances from visual observations~\citep{ha2022flingbot,canberk2023cloth,Wu_2024_CVPR,xue2023unifolding, sunil2025reactiveinairclothingmanipulation}. These structured action spaces improve stability and data efficiency but typically decompose tasks into sequential, open-loop executions with limited adaptability. Complementary approaches learn policies without discrete primitives, including flow-based and model-predictive methods~\citep{weng2022fabricflownet,hoque2020visuospatial,longhini2024adafold}, which improve flexibility and closed-loop control but rely on high-dimensional observations or task-specific geometric representations, limiting the use of compact state abstractions for downstream policy learning. In this work, we train an in-context imitation learning policy in simulation and deploy it zero-shot in the real world.

\paragraph{In-Context Imitation Learning (ICIL).}
ICIL enables task adaptation by conditioning on one or a few demonstrations at test time, without gradient-based fine-tuning. Existing approaches achieve this via explicit alignment between demonstrated and current observations~\citep{vosylius2023few,di2024dinobot,zhang2024oneshot, Wang2025OneShotDI}, meta-learning methods that infer task context from demonstrations~\citep{duan2017one}, and foundation-model-based approaches using structured representations~\citep{dipalo2024kat}. More recently, ICIL has been formulated as next-token prediction conditioned on demonstrations~\citep{fu2024context,zhang2025dynamics}, as well as diffusion-based graph generation~\citep{vosylius2024instant}, enabling more scalable and expressive conditioning for manipulation \citep{qian2026synthicl}. However, these methods focus on rigid-body settings, where object states can often be approximated by low-dimensional pose representations. In contrast, deformable manipulation requires reasoning over high-dimensional, continuously evolving observations and manipulation procedures. Our work extends ICIL to this setting with a novel architecture for long-horizon dual-arm deformable manipulation.

\section{Method}
\label{sec:method}

\begin{figure}[h!]
    \centering
    \includegraphics[width=0.885\linewidth]{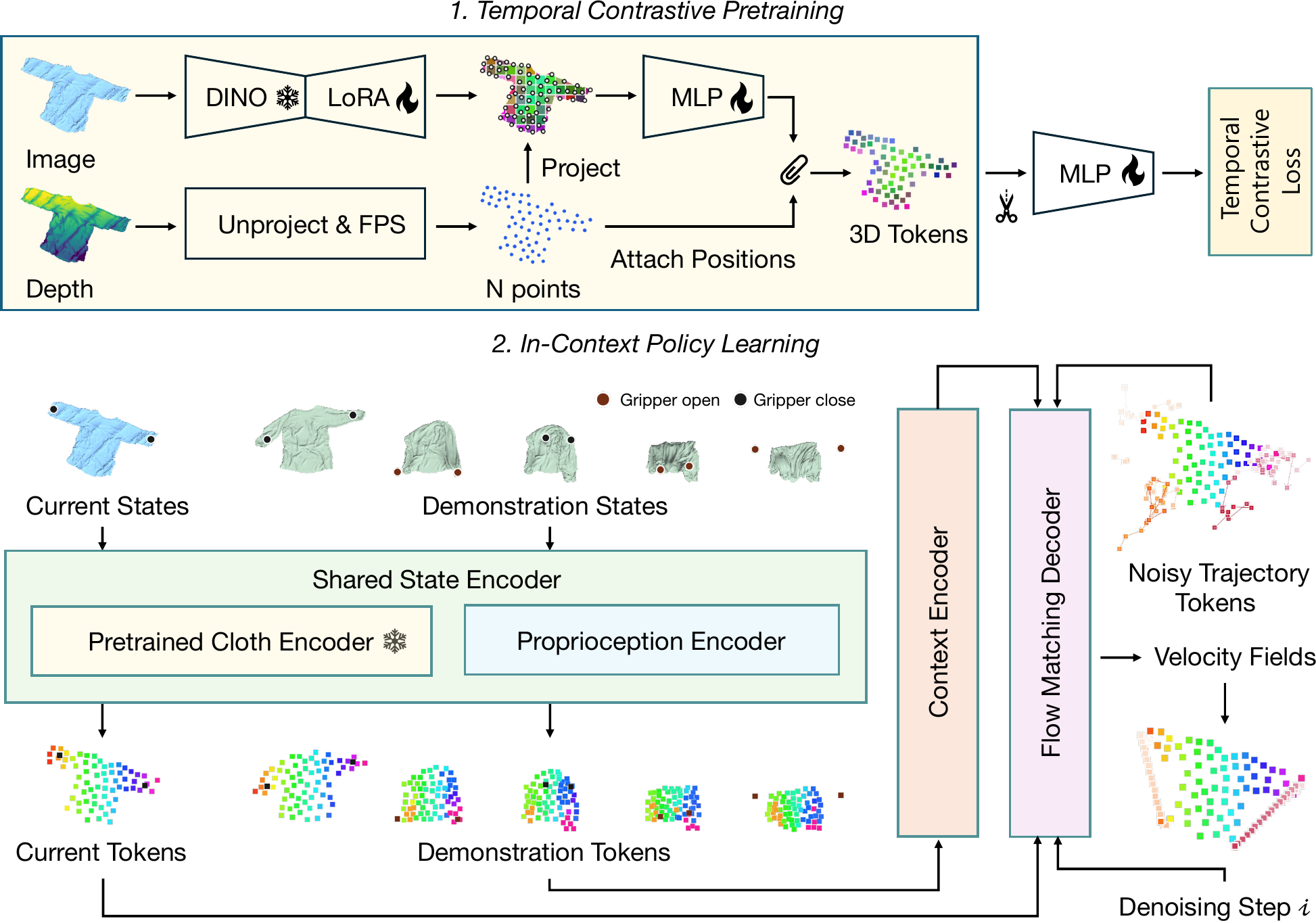}
    \caption{Overview of \textsc{Instant-Fold}. (1) We first pretrain a deformation-aware tokenizer with temporal contrastive supervision, lifting masked RGB-D observations into geo-semantic 3D cloth tokens. (2) During policy learning, the shared encoders tokenize both the demonstration and current observations, aggregate dense spatio-temporal demonstration context, and condition a flow-matching denoising transformer to reconstruct clean dual-arm action trajectories from noise.}
    \label{fig:overview}
\end{figure}

Figure~\ref{fig:overview} provides an overview of our method. \textsc{Instant-Fold} consists of two stages: deformation-aware temporal contrastive pretraining, which learns correspondence-preserving visual tokens, and in-context policy learning, which conditions action generation on demonstration context. Intuitively, the pretrained tokenizer captures how cloth geometry and appearance evolve under deformation, the context encoder infers the intended manipulation procedure from demonstrations, and the action decoder predicts how both arms should move to realize that procedure from the current state.

\paragraph{Deformable Representation.}
We adopt a geometry-first representation for deformable observations. Given a masked RGB-D observation, we back-project valid depth pixels into a 3D point cloud in the camera frame and use Farthest Point Sampling (FPS) to select $N$ points that cover the garment surface. These sampled 3D points are projected back to the image plane, and visual features are extracted from the pretrained image encoder at the projected pixel locations using bilinear interpolation. Each cloth token consists of a 3D position together with its associated semantic feature.

\subsection{Temporal Contrastive Pretraining}
Effective deformable manipulation requires a visual representation that is sensitive to topological state changes, such as folding and crumpling, while remaining robust to nuisance variation such as texture and lighting. We therefore introduce \emph{Temporal Contrastive Pretraining}, a self-supervised objective that leverages dense particle-based simulator geometry \cite{Mller2007PositionBD} to learn a deformation-aware visual encoder from the same simulated trajectory corpus used for downstream policy learning.

\begin{figure}[h]
  \centering
  \begin{subfigure}{0.49\linewidth}
    \centering
    \includegraphics[width=\linewidth]{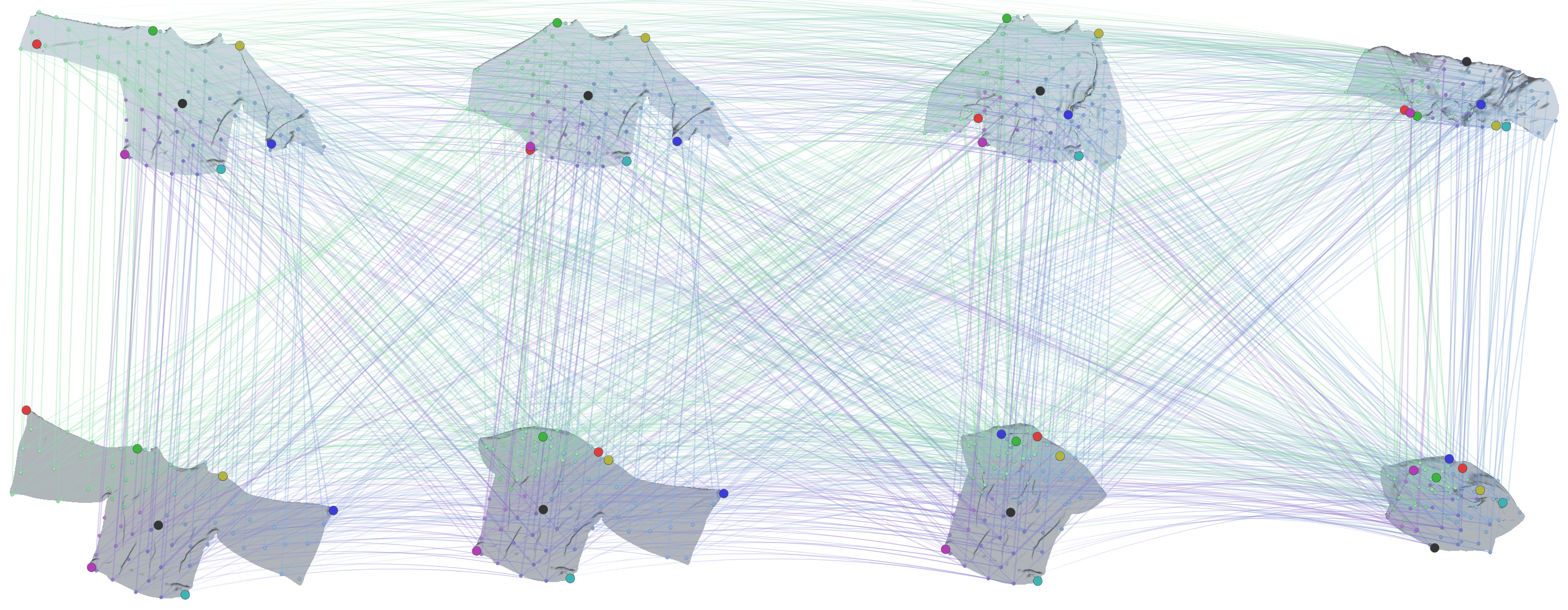}
    \caption*{(a) Same-cloth geometric matching.}
    \label{fig:left_token}
  \end{subfigure}
  \hfill
  \begin{subfigure}{0.49\linewidth}
    \centering
    \includegraphics[width=\linewidth]{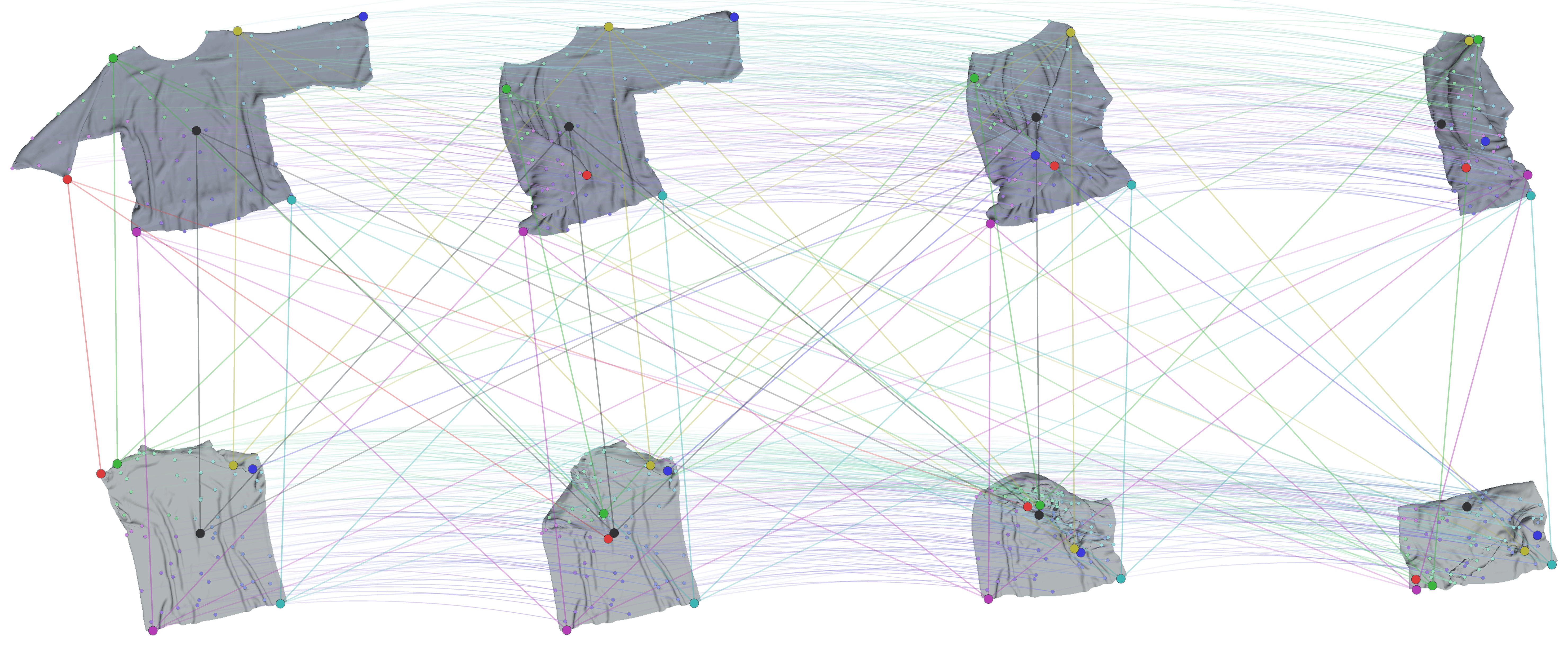}
    \caption*{(b) Cross-cloth semantic matching.}
    \label{fig:right_token}
  \end{subfigure}
  \caption{Illustration of Temporal Contrastive Pretraining with a batch size of two trajectories. (a) \textbf{Same-cloth matching} uses nearest-particle assignment to provide dense correspondence supervision across frames and trajectories of the same cloth. (b) \textbf{Cross-cloth matching} uses sparse semantic keypoints for supervision across different cloth instances. We optimize a soft-weighted contrastive objective with dense negatives from other valid tokens in the comparison set.}
  \vspace{-10pt}
  \label{fig:token_comparison}
\end{figure}

As shown in Figure~\ref{fig:token_comparison}, to ground visual features in physical dynamics, we use particle geometry to define correspondence targets for visual tokens. Our goal is to keep tokens corresponding to the same physical cloth location close in feature space under non-rigid deformation across time.

\paragraph{Loss Function.}
Let $u=(m,t,i)$ index token $i$ at frame $t$ in trajectory $m$, and let $\mathbf{z}_u\in\mathbb{R}^D$ denote its $\ell_2$-normalized feature. For a set of trajectories $\mathcal{S}$ sharing the same cloth identity, let $\pi(u)$ denote the nearest visible simulator particle assigned to token $u$ in 3D. Positives for anchor $u$ are tokens from the same cloth-identity pool that are assigned to the same particle,
\begin{equation}
\mathcal{P}(u)=\{v\neq u:\pi(v)=\pi(u)\}.
\end{equation}
We propose a soft-weighted supervised contrastive loss \cite{khosla2020supervised},
\begin{equation}
\mathcal{L}_{\mathrm{same}}
=
-\frac{1}{|\mathcal{A}|}
\sum_{u\in\mathcal{A}}
\sum_{v\in\mathcal{P}(u)}
\bar{w}_{u,v}
\log
\frac{\exp(\mathbf{z}_u^\top \mathbf{z}_v/\tau)}
{\sum_{v'\in\mathcal{C}(u)}
\exp(\mathbf{z}_u^\top \mathbf{z}_{v'}/\tau)}.
\label{eq:tpc_loss}
\end{equation}
Here, $\mathcal{A}$ contains anchors with at least one visible positive, $\tau$ is the temperature, and $\bar{w}_{u,v}$ are normalized positive weights derived from mutual visibility and Gaussian token-to-particle assignment confidence. The comparison set $\mathcal{C}(u)$ contains valid non-self tokens in the denominator.

To promote transfer across garments, we add a cross-cloth contrastive term using sparse semantic keypoints. Since particle identities are not shared across garments, keypoint labels provide cross-instance correspondence. We form each keypoint feature from nearby visual tokens, treat matching keypoints across garments as positives, and use dense in-batch tokens from other garments as negatives. The full objective is hence $\mathcal{L}_{\mathrm{pretrain}}=\frac{1}{2}\mathcal{L}_{\mathrm{same}}+\frac{1}{2}\mathcal{L}_{\mathrm{cross}}$. More details are in Appendix~\ref{appendix_pretraining}.

\paragraph{Encoder Training.}
We instantiate the encoder with a frozen DINOv3 backbone \cite{simeoni2025dinov3}, aggregate features from multiple transformer depths, adapt the final blocks with LoRA \cite{hu2022lora}, and train a lightweight projection and contrastive head during pretraining. To stabilize optimization and guarantee same-cloth positives within each batch, we use strict PK sampling \cite{hermans2017defense}, where each batch contains $P$ cloth identities and $K$ trajectories per cloth. After pretraining, the contrastive head is discarded, while the LoRA adapters and projection module are transferred to policy learning.

\subsection{In-Context Policy Learning}
We cast DOM as an in-context trajectory generation problem. Given the
current observation $\mathbf{o}_t=(\mathbf{v}_t,\mathbf{q}_t)\in
\mathbb{R}^{d_v}\times\mathbb{R}^{d_q}$, where $\mathbf{v}_t$ denotes
the visual state of the deformable object and $\mathbf{q}_t$ the robot
proprioception, and a context keyframe sequence
\(
\mathcal{C}=(\mathbf{c}_1,\ldots,\mathbf{c}_K), \quad
\mathbf{c}_k=(\mathbf{v}^{\mathrm{demo}}_k,\mathbf{q}^{\mathrm{demo}}_k)
\in\mathbb{R}^{d_v}\times\mathbb{R}^{d_q},
\)
extracted from a single demonstration, the model parameterizes the
conditional distribution
\(
p_\theta\!\left(\mathbf{a}_{t:t+H-1}\mid\mathbf{o}_t,\mathcal{C}\right),
\)
where $\mathbf{a}_{t:t+H-1}\in\mathbb{R}^{H\times d_a}$ is the action trajectory over horizon $H$.

\paragraph{Context Keyframes.}
We automatically extract context keyframes from a single demonstration using end-effector state events, such as binary gripper open/close transitions, following prior work~\cite{james2022q,gkanatsios20253d}. These keyframes summarize temporally distinct object deformation phases while preserving the keypose robot interaction structure that disambiguates the intended folding mode.

\paragraph{Action Representation.}
Each robot state is represented by its Cartesian position
$(x,y,z)\in\mathbb{R}^3$ and binary openness
$o\in\{0,1\}$, with per-gripper action
$\Delta\mathbf{a}^{g}_t=(\Delta x,\Delta y,\Delta z,o)\in
\mathbb{R}^3\times\{0,1\}$ for $g\in\{L,R\}$, giving the full
bimanual action $\mathbf{a}_t=(\Delta\mathbf{a}^{L}_t,
\Delta\mathbf{a}^{R}_t)\in\mathbb{R}^8$. States and actions are represented in the camera frame following~\cite{wang2025observer}, which serves as a shared reference across both arms.

\paragraph{3D Relative Attention.}
Robot state and noisy action tokens are embedded by separate projections while retaining their 3D coordinates. Since cloth, robot states, and noisy action tokens share a common camera-centric 3D frame, we apply relative attention~\cite{gkanatsios20253d} in the geometry-bearing attention blocks using RoPE~\cite{su2024roformer}, allowing attention logits to encode relative positional information.

\paragraph{Context Encoder.}
The context encoder maps the context keyframes $\mathcal{C}$ into a structured representation of intermediate geometry and bimanual interaction. Each keyframe is lifted into cloth tokens via a shared frozen visual tokenizer and concatenated with keypose robot-state tokens. A spatial-then-temporal backbone first applies per-frame spatial attention to model local cloth--robot interactions, then mixes tokens across all keyframes with dense spatio-temporal self-attention. We introduce three context encoding mechanisms: \emph{(i)} 3D
ALiBI~\cite{presstrain} biases spatial attention toward nearby
cloth--robot pairs, grounding cross-modal interactions in 3D geometry; \emph{(ii)} learned summary tokens distill the demonstration into compact global tokens, reducing the decoder burden of inferring long-horizon intent from many low-level tokens; and \emph{(iii)} state-event tokens preserve end-effector gripper events, preventing sparse grasp and release cues from being diluted by the larger cloth-token representation.

\paragraph{Flow Matching Action Decoder.}
The decoder first fuses encoded demonstration context and
current-scene tokens through cross-attention, producing
demonstration-conditioned scene tokens. Noisy trajectory tokens then cross-attend to the scene context formed by concatenating these scene tokens with state-event tokens, and are further refined by joint self-attention. The denoising timestep and current proprioceptive state are encoded by separate MLPs and injected via AdaLN~\cite{peebles2022dit}, conditioning the decoder on both diffusion time and the current bimanual state. An auxiliary keypose prediction task prepends noisy next-keypose tokens to the noisy trajectory tokens; these tokens are decoded with separate position and gripper heads and supervised as short-horizon subgoals.

We train the decoder with rectified flow matching~\cite{liuflow} over
Cartesian positions and binary cross-entropy over gripper openness.
Let $\mathbf{x}$ denote the expert Cartesian trajectory and
$\boldsymbol{\epsilon}\sim\mathcal{N}(0,I)$. We sample $t\in[0,1]$
and form the interpolant
$\mathbf{z}_t=(1-t)\mathbf{x}+t\boldsymbol{\epsilon}$.
The training objective is
\begin{equation}
\mathcal{L}
=
\mathbb{E}_{t,\mathbf{x},\boldsymbol{\epsilon},\mathcal{C}}
\!\left[
\left\|
v_\theta(\mathbf{z}_t,t\mid\mathbf{o}_t,\mathcal{C})
-(\boldsymbol{\epsilon}-\mathbf{x})
\right\|_2^2
+
\operatorname{BCE}(\hat{\mathbf{g}},\mathbf{g})
\right]
+\lambda_{\mathrm{kp}}\mathcal{L}_{\mathrm{kp}},
\end{equation}
where $\hat{\mathbf{g}}$ are predicted gripper-openness logits,
$\mathbf{g}$ are ground-truth gripper labels, and
$\mathcal{L}_{\mathrm{kp}}$ applies the same losses to the auxiliary
keypose prediction. Architecture details are in
Appendix~\ref{architecture_details}.

\section{Experiments}
\label{sec:experiments}
We evaluate \textsc{Instant-Fold} in simulation and on a real dual-arm robot.
Our experiments address six questions:
(1) Does temporal contrastive pretraining improve deformable object representation quality?
(2) If so, does it improve downstream in-context policy learning?
(3) For in-context clothes folding, is demonstration conditioning more effective than language conditioning?
(4) Are our design choices effective?
(5) How does performance scale with context diversity?
(6) How does the final system compare with prior clothes folding methods in simulation and on real garments?

\paragraph{Data Generation.}
\label{sec:data_generation}
We generate demonstrations in a FleX-based simulator~\citep{10.1145/2601097.2601152} with a top-down RGB-D camera and 360 Cloth3D top meshes~\citep{bertiche2020cloth3d}, split into 300 training and 60 held-out meshes. Each trajectory is defined by a folding context: sleeve primitive, body-fold primitive, and execution order. We define 19 modes with valid order variants (Figure~\ref{fig:strategy_bank}) and generate 12 trajectories per mesh per mode, evenly split across variants, yielding ~4K trajectories per mode and ~80K total. Domain randomization covers camera, appearance, initial gripper poses, cloth drop, sleeve disturbances, and wrinkle-like noise fields. For representation pretraining, we add about 40K pretraining-only trajectories from 
semantic-random deformation modes that drag garment regions without defining a folding task. Our main policy benchmark trains on 8 modes (modes 1--6, 9, 11) and evaluates on both seen folds and held-out folds (modes 7, 8, 10). More details are in Appendix~\ref{appendix_data_generation}.

\subsection{Representation Pretraining}
\label{sec:pretrain_eval}
\begin{figure*}[t]
    \centering
    \begin{minipage}[t]{0.49\textwidth}
        \centering
        \includegraphics[width=\linewidth]{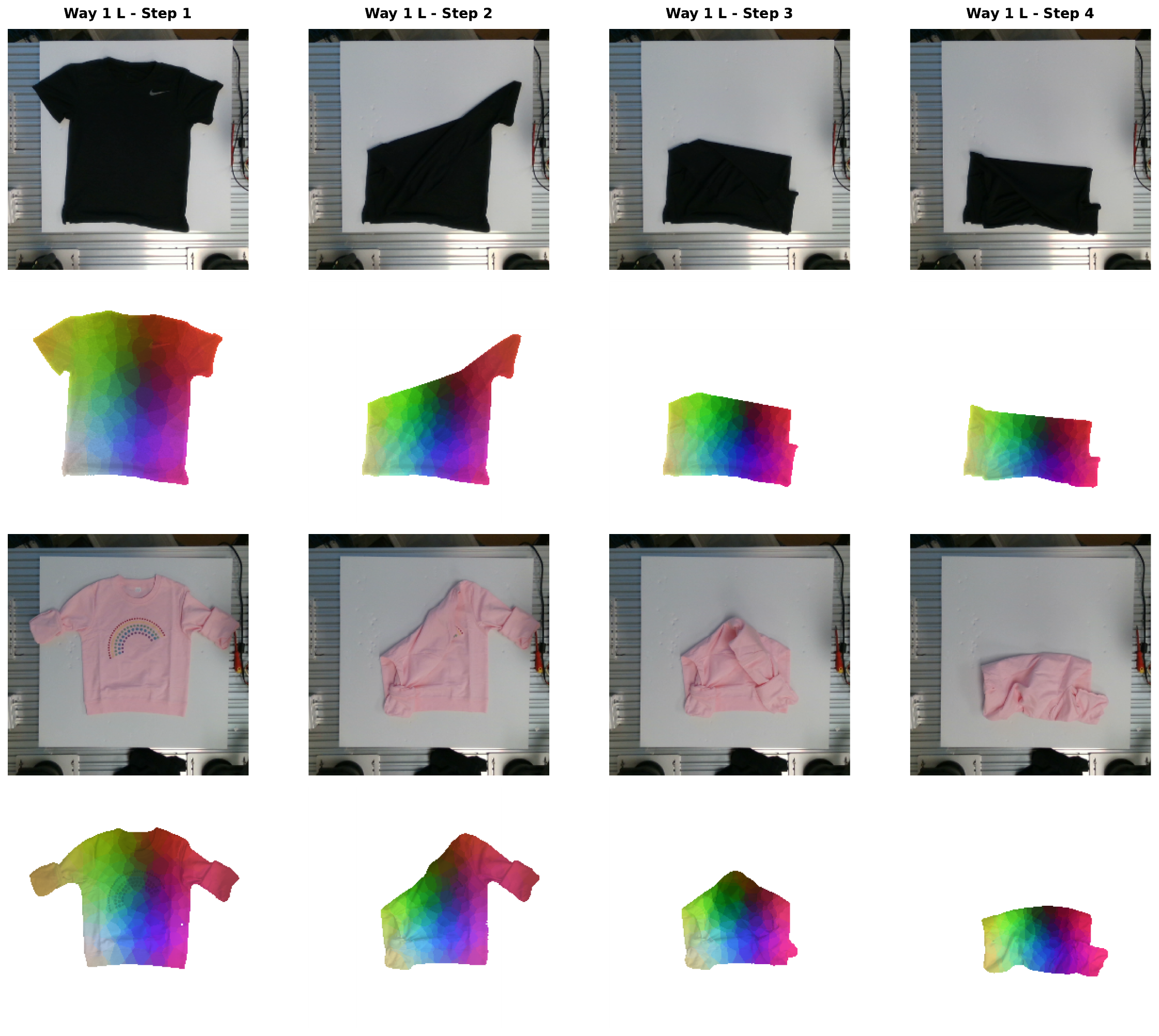}
    \end{minipage}\hfill
    \begin{minipage}[t]{0.49\textwidth}
        \centering
        \includegraphics[width=\linewidth]{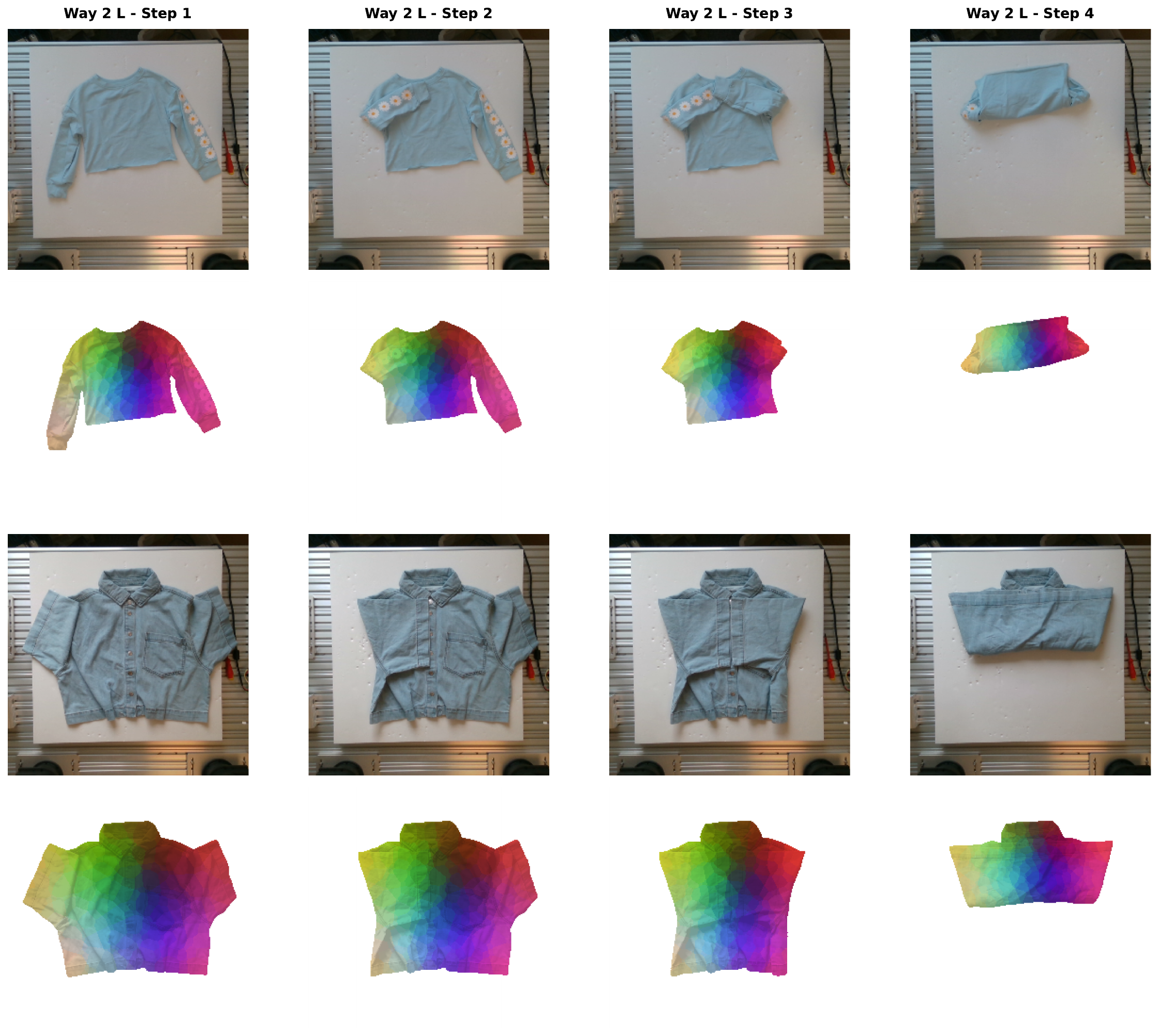}
    \end{minipage}\hfill
    \caption{Visualizations of our pretrained cloth encoder on real-world demonstrations. A shared PCA basis is fit to the token features and projected back onto clothes. For comparison, see Figure~\ref{fig:realworld_pca_triptych_compare}.}
    \vspace{-7pt}
    \label{fig:realworld_pca_triptych}
\end{figure*}

\paragraph{Preprocessing.} For each trajectory, we first filter out static cloth frames by requiring that at least 15\% of cloth particles move more than 2\,cm between consecutive frames. We then uniformly subsample overly long sequences to a fixed frame budget of 15 to maintain realistic VRAM usage.

\paragraph{Training Details.} We use FPS to sample N=64 points. During training, we load only the retained frame indices. We train on a single RTX PRO 6000 Blackwell, using LoRA with rank 32 on the query and value projections of the last 4 transformer blocks of the DINOv3-ViT-S/16+ backbone, together with 4-layer feature aggregation from blocks {3,6,9,12}. Each optimization step uses P=12 cloths and K=8 trajectories per cloth, yielding an effective batch of 96 trajectories. This corresponds to up to 1,440 RGB-D frames (92,160 cloth tokens) per step, with average VRAM of 73.3\,GB.

\paragraph{Results.} As detailed in Appendix~\ref{appendix_pretraining}, our dense-negative temporal formulation with the soft objective consistently achieves the strongest overall performance among the pretraining variants. Figure~\ref{fig:realworld_pca_triptych} shows that the learned encoder transfers effectively from simulation to real-world settings. By comparison (Figure~\ref{fig:realworld_pca_triptych_compare}), DINOv3 produces markedly less temporally consistent representations, while our method yields representations that are both more temporally consistent and spatially coherent than those produced by the pairwise contrastive formulation of ~\cite{Wu_2024_CVPR}, even without real-world adaptation.

\subsection{Policy Learning}
\label{sec:policy_main}
\paragraph{Conditioning variants and baselines.}
We compare two task-conditioning ICL interfaces---language and demonstrations---that both enable zero-shot specification of unseen folding modes without retraining. For the language-conditioned variants, we replace our context encoder with a CLIP \citep{radford2021learning} language encoder, resulting in a policy architecture similar to 3DFA \citep{gkanatsios20253d}, but augmented with our geo-semantic 3D cloth tokens. We evaluate language- and demonstration-conditioned policies under different representation pretraining settings. Without representation pretraining, the policy uses frozen DINOv3 features; with 8-mode pretraining, it is pretrained on the same 8 policy-training modes; and with full pretraining, it is pretrained on 16 folding modes plus additional random deformation modes. To isolate the contribution of our proposed representation, we additionally evaluate the pairwise-correspondence-pretrained PointNet++ \citep{qi2017pointnet++} representation from UniGarment \citep{Wu_2024_CVPR}. To isolate the contribution of our policy architecture, we additionally evaluate ICRT~\citep{11128272}, an autoregressive demonstration-conditioned baseline via causal transformers, using our pretrained visual encoder. The held-out policy contexts are excluded from both pretraining and policy training.

\paragraph{Metrics.}
For each rollout, we generate 20 oracle trajectories with the same initialization as references. We report Ctx.\ Acc., C-SR@95, Geom., and $W_1$. Ctx.\ Acc.\ measures context-following accuracy using an oracle-trained folding-context classifier. C-SR@95 measures conditional fold success: a rollout must both follow the requested folding context and achieve final fold quality within the oracle-calibrated 95th-percentile success envelope. Geom.\ is semantic geometric completion error, computed from context-specific keypoint distances normalized by garment scale. $W_1$ is the Wasserstein-1 distance between rollout and oracle final-state geometry distributions (Appendix~\ref{appendix_evaluation}).

\begin{table}[t]
\centering
\scriptsize
\caption{\textbf{Policy results on 60 held-out garments over 32 folding contexts (1920 rollouts).}
Ctx. Acc. is context-following accuracy. C-SR@95 is conditional success: a rollout must follow the requested context and satisfy the oracle-calibrated 95th-percentile geometric success threshold. Geom. is semantic geometric completion error, and $W_1$ is the Wasserstein-1 distance to the oracle final state.}
\label{tab:policy_main_strict}

\resizebox{\columnwidth}{!}{
\begin{tabular}{llcccccccc}
\toprule
\textbf{Cond.} & \textbf{Pretrain} &
\multicolumn{4}{c}{\textbf{Seen Folds}} &
\multicolumn{4}{c}{\textbf{Held-out Folds}} \\
\cmidrule(lr){3-6} \cmidrule(l){7-10}

& &
Ctx. Acc. $\uparrow$ & C-SR@95 $\uparrow$ & Geom. $\downarrow$ & $W_1$ $\downarrow$ &
Ctx. Acc. $\uparrow$ & C-SR@95 $\uparrow$ & Geom. $\downarrow$ & $W_1$ $\downarrow$ \\
\midrule

Lang. & DINOv3
& 92.2 $\pm$ 2.2
& 35.3 $\pm$ 5.5
& 2.26 $\pm$ 0.21
& .157
& 75.0 $\pm$ 2.6
& 20.2 $\pm$ 4.0
& 3.36 $\pm$ 1.38
& .280 \\

Lang. & Full
& \cellcolor{cvprbest}\textbf{99.0 $\pm$ 0.6}
& 40.4 $\pm$ 5.5
& 2.05 $\pm$ 0.18
& .108
& \cellcolor{cvprsecond}92.0 $\pm$ 2.3
& 28.7 $\pm$ 4.4
& 2.97 $\pm$ 1.24
& .200 \\

\addlinespace[2pt]

Demo & DINOv3
& 91.1 $\pm$ 1.3
& 59.5 $\pm$ 3.9
& 1.57 $\pm$ 0.11
& .057
& 73.5 $\pm$ 2.0
& 42.3 $\pm$ 4.2
& 2.24 $\pm$ 1.01
& .140 \\

Demo & UniGarment
& 98.9 $\pm$ 0.7
& 62.5 $\pm$ 4.1
& 1.49 $\pm$ 0.11
& .046
& 86.5 $\pm$ 2.9
& 23.0 $\pm$ 4.2
& 3.65 $\pm$ 1.42
& .326 \\

Demo & 8-mode
& 98.6 $\pm$ 0.8
& \cellcolor{cvprsecond}68.2 $\pm$ 4.1
& \cellcolor{cvprsecond}1.41 $\pm$ 0.09
& \cellcolor{cvprsecond}.045
& 91.8 $\pm$ 2.5
& \cellcolor{cvprsecond}43.5 $\pm$ 5.3
& \cellcolor{cvprsecond}2.13 $\pm$ 0.60
& \cellcolor{cvprsecond}.137 \\

Demo$^{\mathrm{ICRT}}$ & Full
& 98.5 $\pm$ 0.6
& 49.5 $\pm$ 4.4
& 1.77 $\pm$ 0.12
& .083
& 90.5 $\pm$ 2.4
& 21.3 $\pm$ 3.5
& 3.70 $\pm$ 2.08
& .269 \\

Demo & Full
& \cellcolor{cvprsecond}98.9 $\pm$ 0.7
& \cellcolor{cvprbest}\textbf{69.8 $\pm$ 3.3}
& \cellcolor{cvprbest}\textbf{1.39 $\pm$ 0.08}
& \cellcolor{cvprbest}\textbf{.041}
& \cellcolor{cvprbest}\textbf{95.8 $\pm$ 2.0}
& \cellcolor{cvprbest}\textbf{58.3 $\pm$ 5.0}
& \cellcolor{cvprbest}\textbf{1.89 $\pm$ 0.69}
& \cellcolor{cvprbest}\textbf{.099} \\

\addlinespace[2pt]

Oracle & N/A
& 100.0 $\pm$ 0.0
& 95.0 $\pm$ 0.0
& 1.18 $\pm$ 0.02
& .000
& 100.0 $\pm$ 0.0
& 95.0 $\pm$ 0.0
& 1.17 $\pm$ 0.03
& .000 \\

\bottomrule
\end{tabular}
}
\vspace{-10pt}
\end{table}

\paragraph{Results.}
Table~\ref{tab:policy_main_strict} reveals two main trends. \textbf{(1) Demonstration vs.~Language Conditioning.} Demonstration conditioning substantially improves fold execution quality over language conditioning. Without representation pretraining, held-out context-following accuracy is comparable for language and demonstrations (75.0 vs.~73.5), while demonstrations nearly double held-out C-SR@95 (42.3 vs.~20.2) and achieve better geometric accuracy (Geom.: 2.24 vs.~3.36; $W_1$: 0.140 vs.~0.280). Although the fully pretrained language-conditioned policy achieves the second-highest held-out Ctx.~Acc., it consistently underperforms demonstration-conditioned policies in geometric metrics, suggesting that demonstration keyframes provide explicit spatial-temporal subgoals in the same visual modality as the observations, whereas language provides a more abstract task description. For ICRT, although it achieves higher C-SR@95 on seen folds than the language-conditioned baseline (49.5 vs.~40.4), its performance on held-out folds drops to 21.3, below the fully pretrained language-conditioned policy (28.7). This suggests that the autoregressive sensorimotor formulation of ICRT generalizes less effectively to unseen folding modes than our structured flow-matching approach. \textbf{(2) Representation comparison.} Our representation pretraining improves both language- and demonstration-conditioned policies, with particularly large gains on held-out folds. Pretraining on the 8-mode context subset already achieves the second-best held-out C-SR@95, Geom., and $W_1$ among the demonstration-conditioned policies. Full pretraining yields the strongest overall performance, with the demonstration-conditioned policy achieving 95.8 Ctx.~Acc., 58.3 C-SR@95, 1.89 Geom., and 0.099 $W_1$ on held-out folds, suggesting that broader pretraining coverage produces a more useful representation for downstream policy learning. By contrast, compared to frozen DINOv3 features, UniGarment's pairwise-correspondence representation achieves comparable context-following accuracy but substantially worse geometric performance on held-out folds.

\paragraph{Policy Ablations.}
\label{sec:ablations}
Table~\ref{tab:policy_ablation} evaluates the contribution of each design choice
and two representative combinations. Every single-component ablation reduces
held-out C-SR@95, with the largest drops caused by removing the keypose
auxiliary objective ($-39.4\%$), the context encoder ($-37.1\%$), and the
summary tokens ($-33.7\%$). Qualitatively, a common failure mode is replacing
the coordinated simultaneous fold with two sequential actions. Jointly
removing the context encoder and keypose auxiliary objective produces the
largest degradation across both metrics, including a $-50.9\%$ drop in held-out
C-SR@95 and a $-56.7\%$ drop in geometric gain, supporting their complementary
roles. In contrast, jointly removing state-event and summary tokens has a
non-additive effect: its C-SR@95 degradation remains comparable to the
individual removals, while its held-out geometric gain decreases more
substantially ($-21.9\%$ versus $-6.6\%$ and $-13.7\%$). This suggests partial
redundancy for context-following success but complementary benefits for
geometric generalization.

\begin{table}[h]
\centering
\small
\setlength{\tabcolsep}{6pt}
\renewcommand{\arraystretch}{0.95}
\caption{Policy ablation results. Values denote changes relative to the best model from Table~\ref{tab:policy_main_strict}.}
\label{tab:policy_ablation}
\begin{tabular}{@{}lcccc@{}}
\toprule
Variant
& \multicolumn{2}{c}{Seen Folds}
& \multicolumn{2}{c}{Held-out Folds} \\
\cmidrule(lr){2-3} \cmidrule(l){4-5}
& $\Delta$C-SR@95
& $\Delta$ Geom.~Gain
& $\Delta$C-SR@95
& $\Delta$ Geom.~Gain \\
\midrule

w/o context encoder
& $-10.7\%$ & $-5.6\%$
& $-37.1\%$ & $-27.7\%$ \\

w/o proprioception in context
& $-14.8\%$ & $-6.2\%$
& $-25.1\%$ & $+2.3\%$ \\

w/o keypose auxiliary
& $-8.4\%$ & $-2.1\%$
& $-39.4\%$ & $-36.9\%$ \\

w/o 3D ALiBI
& $-0.4\%$ & $+0.5\%$
& $-25.4\%$ & $-15.6\%$ \\

w/o state-event tokens
& $-7.9\%$ & $-2.3\%$
& $-28.3\%$ & $-6.6\%$ \\

w/o summary tokens
& $-9.8\%$ & $-3.6\%$
& $-33.7\%$ & $-13.7\%$ \\

\addlinespace[2pt]

w/o context encoder \& keypose
& $-22.5\%$ & $-18.8\%$
& $-50.9\%$ & $-56.7\%$ \\

w/o state-event \& summary tokens
& $-10.3\%$ & $-2.2\%$
& $-29.7\%$ & $-21.9\%$ \\

\bottomrule
\end{tabular}
\end{table}

\paragraph{Scaling with Context Diversity.}
\label{sec:scaling}

\begin{wrapfigure}[13]{r}{0.42\textwidth}
    \centering
    \includegraphics[width=0.42\textwidth]{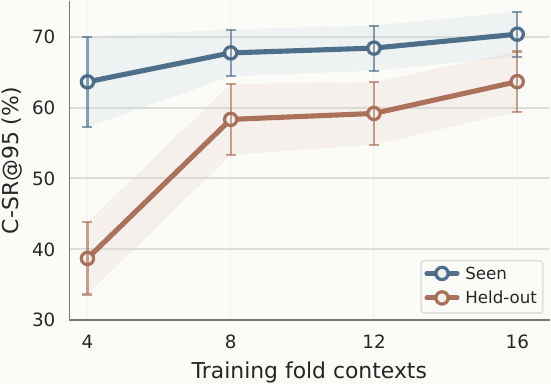}
    \caption{Scaling with context diversity.}
    \label{fig:context_scaling}
\end{wrapfigure}

We vary the context diversity (the number of distinct folding modes available as demonstration contexts during training), while keeping the pretrained encoder and policy recipe fixed. For the seen split, evaluation is restricted to the four seen folds corresponding to modes 1--4 in Figure~\ref{fig:strategy_bank}. Figure~\ref{fig:context_scaling} shows that increasing context diversity primarily improves held-out generalization. Seen-fold C-SR@95 rises gradually from 63.6 with 4 contexts to 70.4 with 16 contexts. In contrast, held-out C-SR@95 improves more substantially, increasing from 38.7 with 4 contexts to 58.3 with 8 contexts and 63.7 with 16 contexts. Notably, the 16-context setting includes four body-first folds (modes 16--19 in Figure~\ref{fig:strategy_bank}); although evaluation remains entirely on sleeve-first folds, these additional body-first contexts still improve generalization. At 16 contexts, the upper tail of held-out performance begins to overlap with the lower tail of seen-fold performance, indicating a substantially reduced generalization gap with broader and more diverse context coverage.

\paragraph{Comparison with Prior Work.}
\label{sec:baselines}
\begin{wraptable}{r}{0.42\textwidth}
\vspace{-8pt}
\centering
\small
\setlength{\tabcolsep}{4pt}

\caption{Results on the two simulators.}
\label{tab:sim_baselines}

\begin{tabular}{@{}lcc@{}}
\toprule
Method & FleX & Isaac Lab \\
\midrule

ClothFunnels
& 71.0 $\pm$ 5.1
& 73.3 $\pm$ 14.8
\\

UniFolding
& \underline{82.0 $\pm$ 4.4}
& \underline{86.7 $\pm$ 10.7}
\\

UniGarment
& 72.7 $\pm$ 5.1
& 83.3 $\pm$ 13.8
\\

\textsc{Instant-Fold}
& \textbf{99.7 $\pm$ 0.7}
& \textbf{92.5 $\pm$ 8.0}
\\

\bottomrule
\end{tabular}

\vspace{-10pt}
\end{wraptable}

For comparison with prior work, we evaluate on unseen meshes and configurations in both FleX (in-distribution, 300 rollouts) and Isaac Lab (sim-to-sim transfer with 2 Franka robots, 120 rollouts) against primitive-based methods: ClothFunnels~\cite{canberk2023cloth}, which combines a keypoint detector with a folding heuristic; UniFolding~\cite{xue2023unifolding}, which predicts action points from point clouds; and UniGarmentManip~\cite{Wu_2024_CVPR}, which estimates action points from a single demonstration via visual correspondence. We evaluate on mode 1, the folding mode shared across all baselines, and adopt the hand-crafted success metric from~\cite{Wu_2024_CVPR}. As shown in Table~\ref{tab:sim_baselines}, our method outperforms all baselines, achieving higher accuracy and lower variance despite operating directly in end-effector space as a closed-loop policy. 

\subsection{Zero-Shot Sim-to-Real with Human Demonstration}
\label{sec:real_world}
We deploy our policy zero-shot on a dual-arm Dobot with a calibrated overhead RGB-D camera. To bridge the robot-occlusion gap, we apply position-dependent gripper occlusions to the current observations during pretraining and policy learning. At inference, we use SAM2~\cite{ravi2025sam} for real-time cloth segmentation; the policy operates on center-cropped RGB-D observations in the camera frame and executes the first 4 of 8 predicted actions at each closed-loop iteration. A human demonstration is provided as only 3--4 keyframe images, with gripper interaction points recovered by a hand tracker or manual labeling, requiring no robot teleoperation. The model performs cloth folding on eight unseen real garments, as shown in Figure 6.~\ref{fig:real_world_clothes}. Additional details are provided in Appendix~\ref{Real-world-appendix}.

\paragraph{Results.}
We quantitatively evaluate Fold~1, the only folding mode shared by all prior simulation-trained cloth-folding baselines, enabling a fair comparison across methods. Table~\ref{tab:real_baselines} reports success rates on eight unseen real garments. Our method achieves higher success across the evaluated garments without requiring real-world data collection or fine-tuning, unlike UniFolding and UniGarmentManip. Additional folding modes, including unseen modes, are demonstrated qualitatively in the accompanying videos at our project page: \url{instant-fold.github.io}.

\begin{figure}[h]
    \centering
    \includegraphics[width=0.995\linewidth]{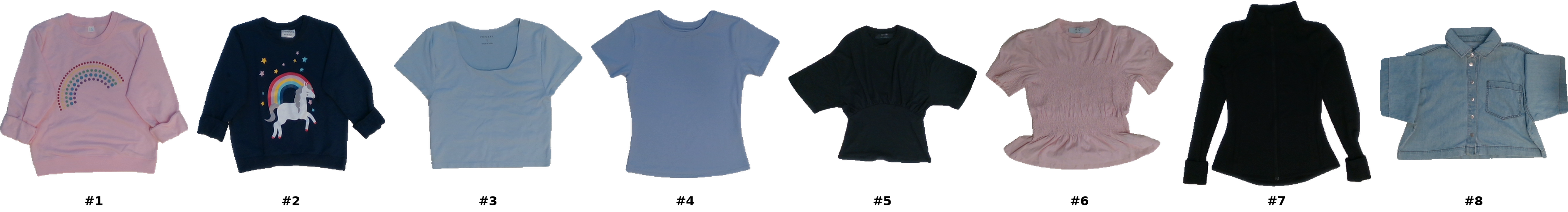}
    \caption{Real-world garments used in our experiments.}
    \label{fig:real_world_clothes}
    \vspace{-10pt}
\end{figure}

\begin{wraptable}{l}{0.58\textwidth}
\vspace{-10pt}
\centering
\small
\setlength{\tabcolsep}{3pt}

\begin{tabular}{@{}lccccccccc@{}}
\toprule
Method 
& \#1 & \#2 & \#3 & \#4 
& \#5 & \#6 & \#7 & \#8 & Avg. \\
\midrule
ClothFunnels
& 3/8 & 1/8 & 2/8 & 0/8 & 0/8 & 0/8 & 0/8 & 0/8 & 9.4 \\
UniFolding
& 5/8 & 4/8 & 3/8 & 2/8 & 3/8 & 1/8 & 1/8 & 0/8 & 29.7 \\
UniGarment
& 2/8 & 4/8 & 2/8 & 2/8 & 2/8 & 1/8 & 3/8 & \textbf{1/8} & 26.6 \\
\textsc{Instant-Fold}
& \textbf{6/8} & \textbf{6/8} & \textbf{6/8} & \textbf{5/8}
& \textbf{6/8} & \textbf{3/8} & \textbf{6/8} & \textbf{1/8} & \textbf{60.9} \\
\bottomrule
\end{tabular}

\caption{Fold~1 success rates on 8 unseen real garments.}
\label{tab:real_baselines}
\vspace{-8pt}
\end{wraptable}

\textbf{Failure Modes.}
Common real-world failure modes include: (1) robot kinematic or controller failures during policy execution, as well as workspace limitations; (2) severe overhead-camera robot occlusions during second-stage folding, causing out-of-distribution observations; (3) sim-to-real physics discrepancies, such as garment edges remaining extended instead of draping downward by gravity, particularly for the stiff garment \#8, or garments sliding during folding on a slippery table; (4) camera calibration drift due to unstable mounting; (5) gripper slippage; and (6) segmentation failures.

\section{Conclusion}
We introduced Instant-Fold, an ICIL framework for DOM in which temporal contrastive pretraining learns deformation-aware visual representations under non-rigid transformations, and an in-context flow-matching policy conditioned on a single demonstration infers and executes the intended manipulation mode at test time. Experiments demonstrate strong performance relative to prior baselines, robust generalization to unseen garments and held-out folding modes, and zero-shot sim-to-real transfer without any real-world training data or fine-tuning. More broadly, our results suggest that ICIL is a promising paradigm for capturing the diversity of valid manipulation modes in DOM, and we hope Instant-Fold can serve as a foundation for broader in-context reasoning in this setting.

\paragraph{Limitations.}
Currently, Instant-Fold has several limitations. First, it assumes reliable object segmentation. Second, our experiments start from states within the foldable manifold rather than from crumpled configurations. Third, robot kinematics and gripper--cloth contact are not modeled during data generation. Fourth, we study only a single object category, namely garment tops. Fifth, our context library is manually designed and limited in scale. A promising future direction is automatic context exploration, which could unleash the full potential of in-context imitation learning and broaden generalization to completely unseen deformable object manipulation tasks.


\clearpage
\acknowledgments{We thank all reviewers and AC for their thorough and insightful feedback to make this paper better.}

\bibliography{example}  

\clearpage
\appendix
\section{Data Generation}
\label{appendix_data_generation}

We generate all training trajectories in the NVIDIA FleX particle simulator \cite{10.1145/2601097.2601152} using scripted dual-arm garment-folding programs executed on 360 long- and short-sleeved top-garment meshes from Cloth3D \cite{bertiche2020cloth3d}. Each rollout records per-frame RGB-D observations (256x256), binary cloth masks, gripper states and actions in the camera frame, camera intrinsics and extrinsics, and simulator geometry in the form of indexed particle positions and visibility masks. To reduce storage, we uniformly downsample each mesh from roughly 12k--28k particles to at most 2k particles.

\paragraph{Trajectory Generation.}
To define semantic manipulation targets, we detect cloth keypoints using a deterministic geometric heuristic on the canonical flattened mesh. After rotating each garment into a standard orientation, we identify corner keypoints as extrema of simple linear functions of particle coordinates, select bottom corners from the lowest hem band, localize one shoulder by detecting the strongest contour-width change and obtain the other by mirror symmetry, and define the center as the particle nearest the centroid of the two shoulders and two bottom corners. We manually inspect the detected keypoints for each mesh and discard degenerate cases. The resulting semantic keypoints are \texttt{top\_left}, \texttt{top\_right}, \texttt{bottom\_left}, \texttt{bottom\_right}, \texttt{left\_shoulder}, \texttt{right\_shoulder}, and \texttt{center}; these keypoints can then be tracked throughout the trajectory by following their associated mesh particles. All folding trajectories are constructed as compositions of keypoint-conditioned pick-and-place subactions followed by gripper retraction. The same generation pipeline is used for both representation pretraining and policy training, with the pretraining set additionally including randomized deformation trajectories for visual pretraining. Each trajectory is generated by initializing a deformable garment in simulation, applying randomized initial-state perturbations, and executing a selected folding program from randomly sampled initial gripper states.

We found the simulator physics to be most stable when scaling each mesh by a factor of two, which increases particle spacing and improves numerical robustness. To preserve a consistent visual scale, we place the camera twice as high and rescale the resulting observations by a factor of one-half.

\paragraph{Context Library and Execution Variants.}
Our folding context library is organized as a collection of semantic folding programs. The main mode families are: \emph{sleeves-first} programs, which first fold the sleeves and then fold the body; \emph{body-first} programs, which reverse this order; and \emph{random-deformation} programs, which apply one or two localized semantic deformations to produce diverse but non-goal-directed cloth configurations for representation pretraining. The sleeves-first family contains 15 modes formed by combining 5 sleeve subactions (\emph{down}, \emph{cross}, \emph{diagonal}, \emph{center}, and \emph{asymmetric}) with 3 body subactions (\emph{shoulders-down}, \emph{bottom-up}, and \emph{side-fold}). In addition, we use 4 body-first modes in our experiments: two \emph{bottom-up} variants and two \emph{side-fold} variants. In our main experiments, we use the 15 sleeves-first modes as folding contexts, and include these 4 body-first contexts only in pretraining and in the 16-context scaling experiments.

To diversify execution while remaining compatible with dual-arm manipulation, each mode is paired with an execution variant that specifies arm ordering. For sleeve subactions, we use left-first (`L`), right-first (`R`), or simultaneous (`S`) execution when safe; for collision-prone sleeve motions, we restrict execution to sequential left-first or right-first variants. For modes containing a directional side fold, a second character specifies the fold direction, yielding variants such as `LL`, `LR`, `RL`, `RR`, `SL`, and `SR`, where the first character denotes sleeve execution order and the second denotes side-fold direction. The low-level controller explicitly avoids gripper interference by using sequential execution and early retraction when pick or place targets would otherwise bring the two arms too close. Figure~\ref{fig:strategy_bank} summarizes these mode families with semantic keyframe visualizations, showing the initial state, an intermediate subfold state, and the final folded state, together with the language descriptions used for language-conditioned policy training and evaluation.

\paragraph{Keyframe Extraction.}
To construct compact demonstration prompts, we extract semantic keyframes from each rollout during trajectory post-processing. We first identify candidate keyframes from gripper-state transitions, including grasp and release events, and always retain the terminal frame. We additionally preserve motion-transition frames that capture sequential coordination patterns, such as one gripper completing its motion before the other begins, which is important for asymmetric and return-early executions. This produces a compact keyframe sequence that preserves the semantic structure of the manipulation while discarding redundant intermediate frames.

\paragraph{Domain Randomization.}
We apply domain randomization to reduce the sim-to-real gap along appearance, sensing, camera, and cloth-state axes. Appearance randomization samples cloth color in HSV space and applies image-space brightness and contrast augmentation. Camera randomization perturbs height, roll, pitch, yaw, and field of view around a real-world-centered operating regime. 

\begin{figure}[t]
    \centering
    \includegraphics[width=0.9\linewidth]{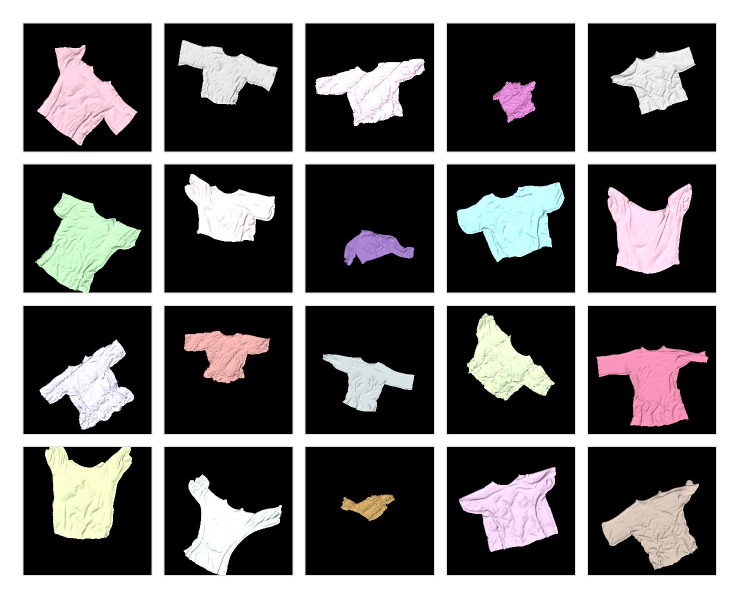}
    \caption{Representative examples of randomized initial cloth configurations used for training, illustrating variation in garment shape, wrinkling, sleeve layout, camera viewpoint, and appearance.
}
    \label{fig:init_configs}
\end{figure}

We also randomize the cloth state itself. First, the garment is spawned with a random in-plane translation and rotation, then dropped from a height of 20\,cm to introduce wrinkles and configuration variation. Second, sleeve disturbances probabilistically perturb the left sleeve, right sleeve, both sleeves, or neither before execution, thereby varying the initial garment layout. Third, a stochastic displacement field is applied with nonzero probability to create additional wrinkles and local nonrigid deformation. In contrast, cloth physics randomization is disabled for simulation stability, so differences across trajectories arise primarily from geometry, initialization, camera, appearance, and cloth-state perturbations rather than changes in simulator material parameters. Table~\ref{tab:domain_randomization} summarizes the full set of randomization parameters. Representative examples of randomized initial cloth configurations are shown in Figure~\ref{fig:init_configs}. We resample if the cloth does not fit within the camera view.

\begin{table}[t]
\centering
\small
\setlength{\tabcolsep}{5pt}
\begin{threeparttable}
\caption{Domain randomization and initial-state variation used for data generation. Unless otherwise noted, parameters are sampled independently and uniformly from the specified ranges.}
\label{tab:domain_randomization}
\begin{tabular}{@{}p{2.6cm}p{3.2cm}p{2.4cm}p{4.7cm}@{}}
\toprule
Category & Parameter & Value / Range & Notes \\
\midrule

\multicolumn{4}{@{}l}{\textit{Cloth layout initialization}} \\
Layout pose & In-plane translation & $[-0.18, 0.18]\,\mathrm{m}$ & Applied along each axis of table plane. \\
Layout pose & In-plane rotation & $[-40^\circ, 40^\circ]$ & Randomized about the table normal. \\
Drop initialization & Cloth drop height & $0.2\,\mathrm{m}$ & Applied before rollout execution. \\

\addlinespace
\multicolumn{4}{@{}l}{\textit{Initial-state variation}} \\
Sleeve layout & Sleeve disturbance mode & \{N, L, R, B\}\tnote{1} & Sampled probabilistically. \\
Sleeve layout & Mode probabilities & $(0.2, 0.2, 0.2, 0.4)$ & Probabilities for \{N, L, R, B\}. \\
Sleeve layout & Displacement amplitude & $[0.04, 0.08]\,\mathrm{m}$ & Applied to the disturbed sleeve configuration. \\
Sleeve layout & Rotation angle & $[-30^\circ, 30^\circ]$ & Applied independently to each disturbed sleeve. \\
Wrinkling & Noise-field probability & $0.5$ & Applied stochastically per trajectory. \\
Wrinkling & Noise-field scale & $[0.9, 1.1]$ & Spatial frequency scaling. \\
Wrinkling & Noise-field amplitude & $[0.02, 0.06]\,\mathrm{m}$ & Magnitude of wrinkle displacement. \\

\addlinespace
\multicolumn{4}{@{}l}{\textit{Appearance}} \\
Cloth color & Hue & $[0.0, 1.0]$ & Sampled uniformly in HSV space. \\
Cloth color & Saturation & $[0.0, 0.8]$ & Includes gray and white garments. \\
Cloth color & Value & $[0.15, 0.9]$ & Includes dark garments. \\
Lighting & Brightness & $[0.9, 1.1]$ & Post-process augmentation. \\
Lighting & Contrast & $[0.95, 1.05]$ & Post-process augmentation. \\

\addlinespace
\multicolumn{4}{@{}l}{\textit{Camera}} \\
Camera pose & Height & $[1.7, 2.3]$ & Simulation-space camera height. \\
Camera pose & Roll, pitch, yaw & $[-5^\circ, 5^\circ]$ & Perturbed around a top-down pose. \\
Camera intrinsics & Field of view & $[35^\circ, 52^\circ]$\tnote{2} & Sampled with a square aspect ratio. \\
Sim-to-real scaling & Target camera height & $1.13\,\mathrm{m}$ & Applied after rendering to match the real setup. \\

\addlinespace
\multicolumn{4}{@{}l}{\textit{Cloth physics}} \\
Material parameters & Stretch, bend, shear & 0.45, 0.05, 0.01 & Physics randomization is disabled for simulation stability. \\
\bottomrule
\end{tabular}

\begin{tablenotes}[flushleft]
\footnotesize
\item[1] N: no sleeve disturbance; L: left-sleeve disturbance; R: right-sleeve disturbance; B: both-sleeve disturbance.
\item[2] The simulator supports only square aspect ratios. During real-world experiments, we center-crop the images.
\end{tablenotes}
\end{threeparttable}
\end{table}

\section{Temporal Contrastive Pretraining}
\label{appendix_pretraining}

We propose a soft-weighted supervised contrastive objective \citep{khosla2020supervised}, where each anchor token may have multiple positives induced by particle correspondence across trajectories and frames sharing the same cloth identity and weighted by a Gaussian function of token-to-particle distance. The same-cloth particle contrastive loss compares tokens within and across trajectories of the same cloth using an all-tokens denominator, and the full algorithm is given in Algorithm~\ref{alg:same_cloth_supcon}. 

\paragraph{Cross-cloth Contrastive Loss.}
To extend correspondence learning across different garment instances, we introduce a cross-cloth objective. Cross-cloth generalization requires learning invariant features that distinguish "same semantic part, different cloth" from "different semantic part, same cloth." This objective reuses the same soft-contrastive machinery as the same-cloth loss, but replaces particle identity with sparse semantic keypoint correspondence across garments. Specifically, each semantic keypoint is encoded by nearby visual tokens, matching keypoints across cloth instances are treated as positives, and all remaining in-batch tokens serve as negatives.

These semantic keypoints are also valid within the same trajectory and across trajectories of the same cloth, but they provide a much sparser supervision signal than dense particle correspondence. We therefore use a curriculum-style weighting over three levels of keypoint matching: same-trajectory, same-cloth cross-trajectory, and cross-cloth, with weights $0.3$, $0.5$, and $1.0$, respectively. This places the strongest emphasis on cross-cloth alignment, which is the most challenging setting. As shown in our experiments, this curriculum weighting performs better than using the cross-cloth term alone.

Implementation-wise, for the same-cloth contrastive loss we found that a fully batch-global denominator, in which each anchor is normalized against all tokens from all trajectories in the batch, was computationally impractical at the batch sizes and token resolutions used in our experiments. This formulation scales quadratically with the total number of batch tokens and requires materializing a dense global similarity matrix over the full batch. We therefore replace the batch-global comparison set with an anchor-wise denominator restricted to tokens from trajectories sharing the same cloth identity. In contrast, for the cross-cloth contrastive loss, a global-denominator formulation remains tractable because the semantic keypoint set is much smaller than the sampled token set (7 vs 64).

\begin{algorithm}[t]
\caption{Same-Cloth Particle Contrastive Loss}
\label{alg:same_cloth_supcon}
\begin{algorithmic}[1]
\REQUIRE A set of trajectories $\mathcal{S}$ sharing the same cloth identity
\REQUIRE Token positions $\mathbf{T}^{(m,t)} \in \mathbb{R}^{N \times 3}$ and features $\mathbf{F}^{(m,t)} \in \mathbb{R}^{N \times D}$
\REQUIRE Particle positions $\mathbf{P}^{(m,t)} \in \mathbb{R}^{M \times 3}$, visibility $\mathbf{v}^{(m,t)} \in \{0,1\}^{M}$
\REQUIRE Gaussian width $\sigma$, temperature $\tau$

\STATE \textbf{Assign tokens to visible particles}
\FOR{$m \in \mathcal{S}$}
    \FOR{$t \in \{0,\ldots,T_m-1\}$}
        \STATE $D^{(m,t)}_{ip} \gets \|\mathbf{T}^{(m,t)}_i - \mathbf{P}^{(m,t)}_p\|$
        \STATE $D^{(m,t)}_{ip} \gets +\infty$ where $\mathbf{v}^{(m,t)}_p = 0$
        \STATE $a^{(m,t)}_i \gets \arg\min_p D^{(m,t)}_{ip}$
        \STATE $d^{(m,t)}_i \gets \|\mathbf{T}^{(m,t)}_i - \mathbf{P}^{(m,t)}_{a^{(m,t)}_i}\|$
    \ENDFOR
\ENDFOR

\STATE \textbf{Collect tokens and define shorthand}
\STATE $\mathcal{I} \gets \{(m,t,i)\mid m\in\mathcal{S},\ t\in\{0,\ldots,T_m-1\},\ i\in\{1,\ldots,N\}\}$
\STATE For $u=(m,t,i)\in\mathcal{I}$, define $a_u \triangleq a_i^{(m,t)}$, $d_u \triangleq d_i^{(m,t)}$, and $\mathbf{z}_u \triangleq \mathbf{F}_i^{(m,t)}/\|\mathbf{F}_i^{(m,t)}\|$.

\STATE \textbf{Build positive weights}
\FOR{$u,v \in \mathcal{I}\times\mathcal{I}$}
    \STATE $M_{u,v} \gets \mathbbm{1}[a_u = a_v]$
    \STATE $W_{u,v} \gets M_{u,v}\exp\!\left(-\frac{d_u^2+d_v^2}{4\sigma^2}\right)$
\ENDFOR
\STATE $W_{u,u} \gets 0$ for all $u\in\mathcal{I}$

\STATE \textbf{Compute soft-weighted supervised contrastive loss}
\FOR{$u\in\mathcal{I}$}
    \STATE $\mathcal{C}(u) \gets \mathcal{I}\setminus\{u\}$
    \STATE $\bar{W}_{u,v} \gets \dfrac{W_{u,v}}{\sum_{v'} W_{u,v'}}$ for $v\in\mathcal{C}(u)$
    \STATE $\ell_u \gets - \sum_{v\in\mathcal{C}(u)} \bar{W}_{u,v}
    \log \dfrac{\exp(\mathbf{z}_u^\top \mathbf{z}_v/\tau)}
    {\sum_{v'\in\mathcal{C}(u)} \exp(\mathbf{z}_u^\top \mathbf{z}_{v'}/\tau)}$
\ENDFOR

\RETURN $\mathrm{mean}\big(\{\ell_u : \sum_v W_{u,v} > 0\}\big)$
\end{algorithmic}
\end{algorithm}

\paragraph{Forward-Only Contrast.}
We use source-to-target correspondence only because cloth folding induces many-to-one mappings under self-occlusion: multiple source tokens may map to the same particle region, whereas the reverse assignment is often ambiguous. Averaging forward and backward objectives can therefore introduce conflicting gradients and destabilize training.

\paragraph{Trajectory Filtering.}
To reduce redundant trajectories, such as when the robot grippers move but the clothes do not, we retain only frames that exhibit sufficient cloth motion. We initialize the kept-frame set as $S=\{0\}$ and then iterate over frames $t=1,\dots,T-1$. Let $t_{\mathrm{last}}=\max(S)$ denote the most recently retained frame. For each particle $k$, we compute its displacement relative to the last kept frame,
\(
\delta_k = \left\| P_t^{(k)} - P_{t_{\mathrm{last}}}^{(k)} \right\|_2.
\)
We retain frame $t$ if the fraction of particles whose displacement exceeds a distance threshold is larger than a preset motion threshold:
\(
\frac{1}{K}\sum_{k=1}^{K}\mathbb{I}\!\left(\delta_k > \tau_{\mathrm{dist}}\right) > \tau_{\mathrm{frac}}.
\)
In all experiments, we use $\tau_{\mathrm{dist}}=0.02\,\mathrm{m}$ and $\tau_{\mathrm{frac}}=0.15$. If a trajectory has more than 15 frames, we uniformly subsample it to 15 frames to keep computation tractable.

\begin{table}[t]
\centering
\caption{Intra-trajectory correspondence tracking. Mean error in centimeters (mean $\pm$ std) and PCK (\%) on the evaluation subset. KF: keyframes only; Filt: motion-filtered trajectories with a 2\,cm threshold. Bold indicates best overall; underlined values indicate the best in each method family.}
\label{tab:intra_traj}
\vspace{0.3em}
\small
\setlength{\tabcolsep}{6pt}
\begin{sc}
\begin{tabular}{@{}llccc@{}}
\toprule
\textbf{Method} & \textbf{Data} & \textbf{Err}$\downarrow$ & \textbf{PCK@5}$\uparrow$ & \textbf{PCK@2}$\uparrow$ \\
\midrule
\multicolumn{5}{@{}l}{\textit{Structured (consecutive + goal-anchored + keyframes)}} \\[2pt]
\quad Soft + dist.\ wt. & KF & 5.97{\scriptsize$\pm$2.5} & 57.7 & 32.6 \\
\quad Soft + dist.\ wt. & Filt & $\underline{5.53}${\scriptsize$\pm$\underline{2.9}} & 57.1 & 32.5 \\
\quad Soft & Filt & 14.6{\scriptsize$\pm$12.1} & 57.1 & 32.2 \\
\quad Soft + Symmetric & KF & 19.0{\scriptsize$\pm$11.3} & \underline{59.2} & \underline{33.3} \\
\midrule
\multicolumn{5}{@{}l}{\textit{Exhaustive (all token$\to$particle pairs)}} \\[2pt]
\quad Soft & KF & 8.15{\scriptsize$\pm$4.6} & 50.5 & 29.0 \\
\quad Soft & Filt & \underline{6.34}{\scriptsize$\pm$3.2} & 49.2 & 28.6 \\
\quad Soft + Symmetric & KF & 26.2{\scriptsize$\pm$11.2} & 53.5 & 31.1 \\
\quad Hard & KF & 10.5{\scriptsize$\pm$6.9} & 53.9 & \underline{31.4} \\
\quad Hard & Filt & 10.6{\scriptsize$\pm$7.8} & \underline{54.2} & \underline{31.4} \\
\midrule
\multicolumn{5}{@{}l}{\textit{Symmetric (outer-product weights)}} \\[2pt]
\quad Soft & KF & 7.69{\scriptsize$\pm$2.4} & \underline{37.8} & \underline{21.6} \\
\quad Soft & Filt & \underline{6.85}{\scriptsize$\pm$1.6} & 33.1 & 20.2 \\
\midrule
\multicolumn{5}{@{}l}{\textit{Temporal Contrast (ours)}} \\[2pt]
\quad Soft + dist.\ wt. & Filt & 6.01{\scriptsize$\pm$2.6} & 59.8 & 33.4 \\
\quad Soft & Filt & \textbf{5.51}{\scriptsize$\pm$\textbf{1.9}} & \textbf{60.6} & \textbf{34.1} \\
\quad Hard & Filt & 10.4{\scriptsize$\pm$11.4} & 55.1 & 32.3 \\
\midrule
DINOv3-S+ (frozen) & --- & 21.2{\scriptsize$\pm$2.0} & 28.0 & 23.1 \\
\bottomrule
\end{tabular}
\end{sc}
\end{table}

\subsection*{Additional Ablation Experiments.}
We present additional representation-pretraining ablations to justify our design choices. For efficiency, these experiments use a subset of the full pretraining corpus and evaluate representation quality using mean tracking error and PCK (Percentage of Correct Keypoints). We compute these metrics with a feature-space $k$-NN correspondence probe: for each query token, we retrieve its nearest neighbors in the target frame by cosine similarity, predict its 3D location by similarity-weighted averaging, and compare it against the ground-truth particle position.

\paragraph{Intra-Trajectory Ablation.}
Table~\ref{tab:intra_traj} compares four loss families, trained with a batch size of 1 and evaluated on intra-trajectory correspondence tracking: structured pairwise contrast, exhaustive pairwise contrast, symmetric pairwise contrast, and our dense temporal contrastive objective. Structured variants use selected temporal relations, including consecutive pairs, goal-anchored pairs that match each frame to future keyframes, and intermediate keyframe pairs. Exhaustive variants match all eligible frame pairs, whereas symmetric variants use bidirectional matching or symmetric outer-product weights. Our method instead applies a supervised contrastive objective over all valid tokens in the trajectory, with positives defined by shared particle assignments and each anchor normalized against all non-self tokens in the trajectory. Across these families, \emph{hard} denotes binary positive assignments induced by nearest-particle correspondences, whereas \emph{soft} denotes Gaussian-weighted positive assignments based on token-to-particle distance. \emph{Distance weighting} further reweights positive pairs by temporal separation, placing greater emphasis on longer-range correspondences. The key findings are:

\begin{enumerate}[leftmargin=*, topsep=0pt, itemsep=3pt]
\item \textbf{Structured pairing outperforms exhaustive pairing.}
Structured variants consistently achieve lower tracking error than exhaustive variants, suggesting that targeted temporal pairs provide more informative supervision than uniformly matching all eligible frame pairs.

\item \textbf{Motion filtering is most helpful for exhaustive objectives.}
Filtering improves exhaustive matching substantially, since low-motion frames contribute redundant or weak supervision. Structured objectives benefit less because they already emphasize informative temporal relations.

\item \textbf{Soft targets are more stable than hard targets.}
Soft Gaussian weighting yields lower error and lower variance than hard assignment, indicating greater robustness to token-sampling noise, nondeterministic FPS, partial occlusion, and correspondence ambiguity in deformable clothes.

\item \textbf{Symmetric losses degrade correspondence quality.}
Symmetric variants increase tracking error substantially, suggesting that bidirectional matching introduces ambiguous supervision under self-occlusion, where a token may have multiple plausible backward matches.

\item \textbf{Our dense formulation performs best.}
Using all valid anchor-positive pairs with dense negatives and soft weighting yields the strongest results, suggesting that a richer contrastive signal helps the model learn features invariant to temporal position, deformation state, and cloth configuration. It also indicates that explicit distance weighting becomes unnecessary when the denominator is sufficiently dense, although optimization is more demanding.
\end{enumerate}

\begin{table}[h]
\centering
\caption{Correspondence tracking results. Mean error (cm) with std and PCK@5cm (\%) across 200 trajectories. \textbf{Bold}: best per column. Seen/Unseen: training vs.\ held-out folding modes.}
\begin{small}
\begin{sc}
\label{tab:supcon_ablation}
\resizebox{\textwidth}{!}{%
\begin{tabular}{@{}l@{\hspace{8pt}}cc@{\hspace{12pt}}cc@{\hspace{16pt}}cc@{\hspace{12pt}}cc@{}}
\toprule
& \multicolumn{4}{c}{\textbf{Seen Fold}} & \multicolumn{4}{c}{\textbf{Unseen Fold}} \\
\cmidrule(lr){2-5} \cmidrule(lr){6-9}
\textbf{Method} & \multicolumn{2}{c}{Intra-Traj} & \multicolumn{2}{c}{Cross-Traj} & \multicolumn{2}{c}{Intra-Traj} & \multicolumn{2}{c}{Cross-Traj} \\
\cmidrule(lr){2-3} \cmidrule(lr){4-5} \cmidrule(lr){6-7} \cmidrule(lr){8-9}
& Err$\downarrow$ & PCK$\uparrow$ & Err$\downarrow$ & PCK$\uparrow$ & Err$\downarrow$ & PCK$\uparrow$ & Err$\downarrow$ & PCK$\uparrow$ \\
\midrule
\multicolumn{9}{l}{\textit{Pairwise Contrast (single-pair denominator)}} \\[2pt]
\quad Structured & 5.88{\scriptsize$\pm$7.9} & 58.5 & 7.15{\scriptsize$\pm$10.2} & 46.5 & 8.41{\scriptsize$\pm$10.8} & 37.8 & 9.63{\scriptsize$\pm$14.6} & 36.5 \\
\quad + Cross-Traj & 6.28{\scriptsize$\pm$10.5} & 60.9 & 5.40{\scriptsize$\pm$8.4} & 64.3 & 9.00{\scriptsize$\pm$10.8} & 38.5 & 8.68{\scriptsize$\pm$15.4} & 46.6 \\
\quad + Cross-Context & 5.76{\scriptsize$\pm$8.7} & 62.3 & 5.01{\scriptsize$\pm$6.4} & 65.0 & 8.72{\scriptsize$\pm$9.8} & 38.1 & 8.33{\scriptsize$\pm$14.9} & 47.7 \\
\midrule
\multicolumn{9}{l}{\textit{Temporal Contrast (our, all-frames denominator)}} \\[2pt]
\quad Soft + Equal weight& 6.12{\scriptsize$\pm$5.8} & 56.7 & 5.53{\scriptsize$\pm$3.7} & 53.0 & 9.04{\scriptsize$\pm$7.8} & 36.4 & 7.23{\scriptsize$\pm$5.5} & 39.3 \\
\quad + Cross-Traj & 4.99{\scriptsize$\pm$3.0} & 60.3 & 4.47{\scriptsize$\pm$2.5} & 65.0 & 8.11{\scriptsize$\pm$6.3} & 37.5 & 6.01{\scriptsize$\pm$3.9} & 48.2 \\
\quad + Cross-Context & \textbf{4.79{\scriptsize$\pm$2.8}} & \textbf{63.2} & \textbf{4.36{\scriptsize$\pm$2.5}} & \textbf{66.3} & \textbf{8.09{\scriptsize$\pm$6.4}} &  \textbf{38.4} & \textbf{5.98{\scriptsize$\pm$4.0}} & \textbf{48.4} \\
\midrule
\multicolumn{9}{l}{\textit{Temporal Contrast Ablations (all with Cross-Context)}} \\[2pt]
\quad w/ Hard assignment & 9.04{\scriptsize$\pm$15.7} & 53.4 & 5.92{\scriptsize$\pm$5.7} & 52.1 & 13.5{\scriptsize$\pm$20.6} & 33.3 & 10.0{\scriptsize$\pm$15.3} & 37.1 \\
\quad w/ Distance weight & 6.52{\scriptsize$\pm$6.7} & 56.2 & 5.47{\scriptsize$\pm$3.7} & 53.1 & 9.75{\scriptsize$\pm$8.7} & 34.3 & 7.34{\scriptsize$\pm$5.9} & 39.4 \\

\bottomrule
\end{tabular}%
}
\end{sc}
\end{small}
\end{table}

\paragraph{Cross-Trajectory Ablation.}
We increase the batch size to 4 to accommodate multiple trajectories from the same cloth identity within a batch and evaluate cross-trajectory and cross-context supervision. Table~\ref{tab:supcon_ablation} compares structured pairwise contrast with our dense temporal contrastive objective on both seen and unseen folding modes. Across both splits, adding cross-trajectory positives improves correspondence matching, and adding cross-context positives provides a further gain. Overall, our dense temporal formulation is consistently strongest. The Key Findings ares:

\begin{enumerate}[leftmargin=*, topsep=0pt, itemsep=3pt]
\item \textbf{Dense temporal contrast performs best overall.}
Compared with structured pairwise contrast, the dense all-frames denominator consistently achieves lower error and higher PCK across all columns, with the largest gains on cross-trajectory matching. This suggests that exposing each anchor to a richer set of positives and harder negatives leads to more discriminative, context-invariant features, and that the same advantage extends naturally across trajectories.

\item \textbf{Cross-trajectory and cross-context supervision are complementary.}
Adding cross-trajectory positives improves performance, and adding cross-context positives yields a further improvement, especially on seen folds. This suggests that cross-trajectory supervision promotes invariance to initialization and deformation, while cross-context supervision further improves consistency across different manipulation modes and cloth configurations.

\item \textbf{Soft dense weighting is preferable to hard or distance-weighted variants.}
Hard assignment and explicit distance weighting both degrade performance, indicating that soft positives with a dense denominator provide the most effective supervision. Soft weighting is more robust to ambiguous correspondences caused by deformation, occlusion, and token-sampling noise, while explicit distance weighting appears unnecessary once the denominator is sufficiently dense.
\end{enumerate}

\begin{table}[t]
\centering
\caption{\textbf{Ablation study on representation learning components.} We report tracking errors (cm, $\downarrow$) and PCK@5cm (\%, $\uparrow$). Relative changes from our full model are shown in parentheses.}
\label{tab:ablation_pretrain}
\vspace{0.5em}
\resizebox{\linewidth}{!}{%
\begin{tabular}{@{}l cc cc cc@{}}
\toprule
& \multicolumn{2}{c}{\textbf{Intra-Traj}} & \multicolumn{2}{c}{\textbf{Cross-Traj}} & \multicolumn{2}{c}{\textbf{Cross-Cloth}} \\
\cmidrule(lr){2-3} \cmidrule(lr){4-5} \cmidrule(lr){6-7}
\textbf{Configuration} & Err $\downarrow$ & PCK $\uparrow$ & Err $\downarrow$ & PCK $\uparrow$ & Err $\downarrow$ & PCK $\uparrow$ \\
\midrule
\textbf{Full Model (Ours)} & \textbf{3.84} & \textbf{75.7} & \textbf{6.18} & \textbf{57.2} & 7.43 & \textbf{52.7} \\
\midrule
w/o Multi-Layer & 4.70 \scriptsize{(+22\%)} & 72.2 \scriptsize{(-5\%)} & 7.18 \scriptsize{(+16\%)} & 51.8 \scriptsize{(-9\%)} & 9.53 \scriptsize{(+28\%)} & 39.1 \scriptsize{(-26\%)} \\
w/o Curriculum Weight & 3.91 \scriptsize{(+2\%)} & 74.3 \scriptsize{(-2\%)} & 6.66 \scriptsize{(+8\%)} & 53.8 \scriptsize{(-6\%)} & 7.54 \scriptsize{(+1\%)} & 49.5 \scriptsize{(-6\%)} \\
w/o Temp. Annealing & 3.86 \scriptsize{(+1\%)} & 74.8 \scriptsize{(-1\%)} & 6.50 \scriptsize{(+5\%)} & 53.7 \scriptsize{(-6\%)} & \textbf{7.36 \scriptsize{(-1\%)}} & 49.5 \scriptsize{(-6\%)} \\
+ H-SCL \cite{robinsoncontrastive} & 4.18 \scriptsize{(+9\%)} & 71.2 \scriptsize{(-6\%)} & 6.71 \scriptsize{(+9\%)} & 52.3 \scriptsize{(-9\%)} & 8.11 \scriptsize{(+9\%)} & 36.4 \scriptsize{(-31\%)} \\
+ FPN Neck & 4.02 \scriptsize{(+5\%)} & 73.7 \scriptsize{(-3\%)} & 6.59 \scriptsize{(+7\%)} & 52.9 \scriptsize{(-8\%)} & 8.90 \scriptsize{(+20\%)} & 44.6 \scriptsize{(-15\%)} \\
\bottomrule
\end{tabular}%
}
\vspace{-1em}
\end{table}

\paragraph{Pretraining Ablations.}
Table~\ref{tab:ablation_pretrain} shows that the full model achieves the best overall balance across intra-trajectory, cross-trajectory, and cross-cloth correspondence. The largest gain comes from multi-layer DINOv3 feature aggregation, which improves all metrics and is especially important for cross-cloth matching, suggesting that semantic correspondence benefits from combining mid-level cues from blocks 3, 6, 9, and 12 rather than relying on the final layer alone. Curriculum weighting is applied to the sparse semantic keypoint contrasting objective, with same-trajectory, same-cloth cross-trajectory, and cross-cloth terms weighted by $0.3/0.5/1.0$; this stabilizes optimization and helps preserve the same-cloth particle-tracking knowledge learned from the dense correspondence loss. Temperature annealing provides a smaller but still measurable stabilization effect by gradually sharpening the contrastive comparisons during training.

By contrast, adding H-SCL \cite{robinsoncontrastive} or an FPN neck does not help in our setting and often hurts performance, especially for cross-cloth correspondence. This suggests that our task benefits more from a clean dense contrastive objective and semantically rich multi-layer features than from additional hard-negative mining or architectural complexity. Overall, the ablation indicates that the main gains come from better feature aggregation and more balanced optimization.

\paragraph{Full Pretraining Results.}
We report the final results from full pretraining on the 60 unseen clothes meshes used for evaluation in Table~\ref{tab:pretrain_blackwell_ablation}. After scaling to 16 contexts and 40K random-deformation trajectories, the model learns intra-trajectory and cross-cloth correspondences well, while cross-trajectory matching remains the main bottleneck. This suggests that the hardest part is learning precise correspondences across the wide range of possible deformations encountered during folding.

\begin{table}[h]
\centering
\caption{\textbf{Final results from our full pretraining runs.} We report tracking error (cm, $\downarrow$), PCK@2cm (\%, $\uparrow$), and PCK@5cm (\%, $\uparrow$).}
\label{tab:pretrain_blackwell_ablation}
\vspace{0.5em}

\resizebox{\linewidth}{!}{%
\begin{tabular}{@{}ccc ccc ccc@{}}
\toprule
\multicolumn{3}{c}{\textbf{Intra-Traj}} &
\multicolumn{3}{c}{\textbf{Cross-Traj}} &
\multicolumn{3}{c}{\textbf{Cross-Cloth}} \\
\cmidrule(lr){1-3} \cmidrule(lr){4-6} \cmidrule(lr){7-9}

Err $\downarrow$ & PCK@2 $\uparrow$ & PCK@5 $\uparrow$
& Err $\downarrow$ & PCK@2 $\uparrow$ & PCK@5 $\uparrow$
& Err $\downarrow$ & PCK@2 $\uparrow$ & PCK@5 $\uparrow$ \\
\midrule

3.60 & 24.8 & 80.6
& 6.21 & 17.6 & 60.8
& 2.72 & 49.1 & 86.7 \\

\bottomrule
\end{tabular}%
}
\end{table}

\begin{figure*}[t]
    \centering
    \begin{minipage}[t]{0.49\textwidth}
        \centering
        \includegraphics[width=\linewidth]{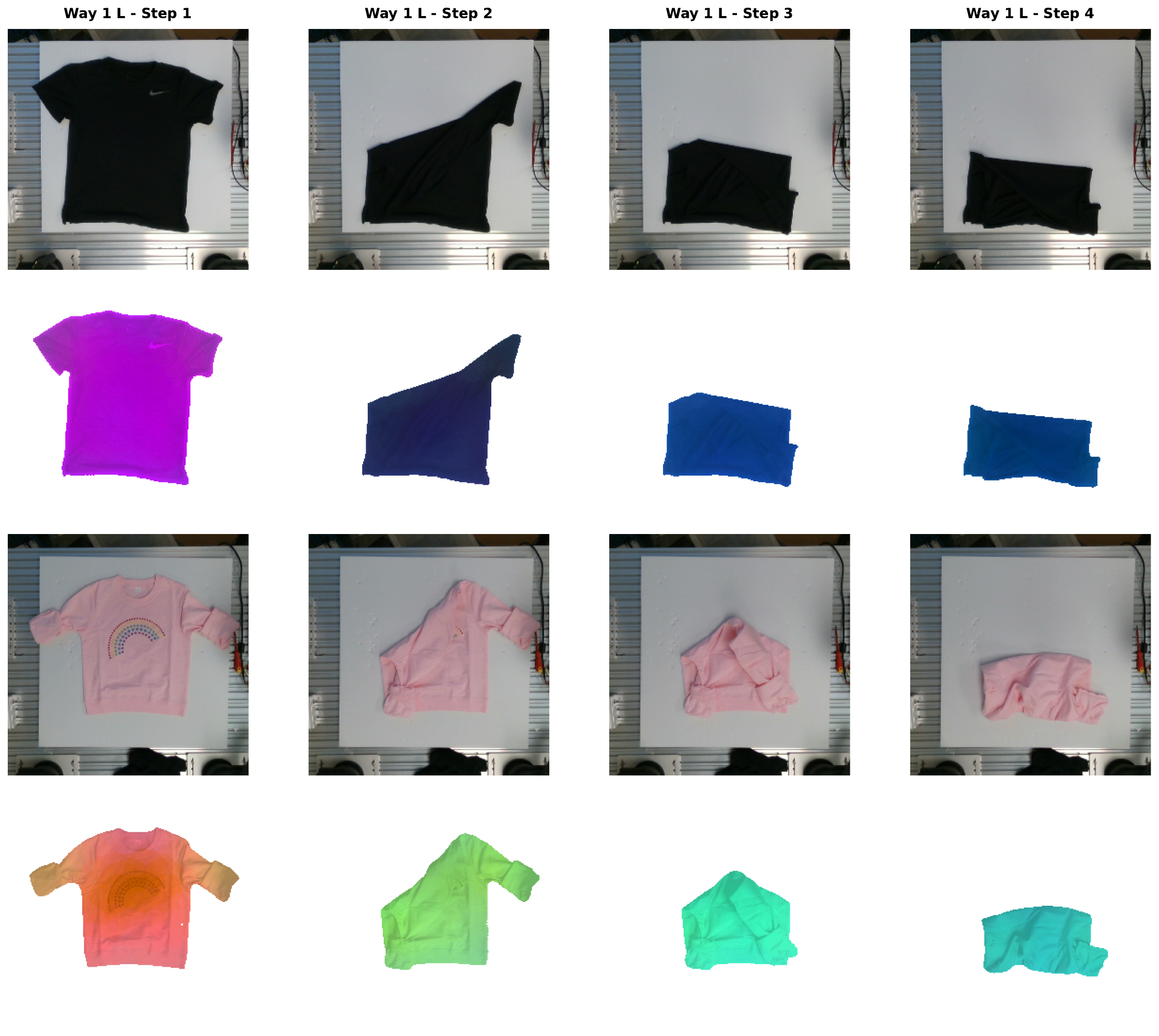}
    \end{minipage}\hfill
    \begin{minipage}[t]{0.49\textwidth}
        \centering
        \includegraphics[width=\linewidth]{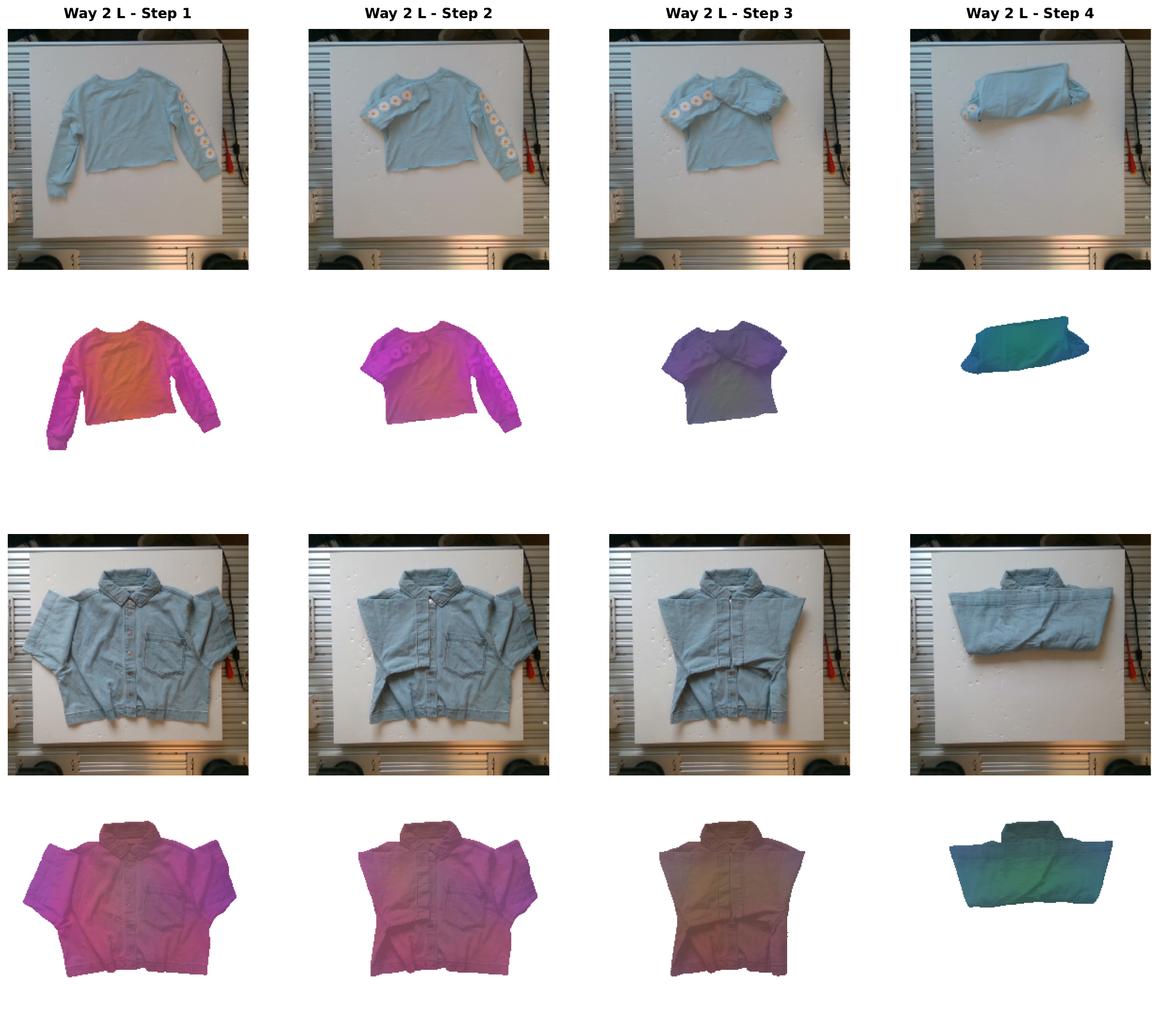}
    \end{minipage}

    \vspace{2mm}
    {\small (a) DINOv3 (ViT-S/16+) \cite{simeoni2025dinov3} representations on real-world demonstrations.}

    \vspace{3mm}

    \begin{minipage}[t]{0.49\textwidth}
        \centering
        \includegraphics[width=\linewidth]{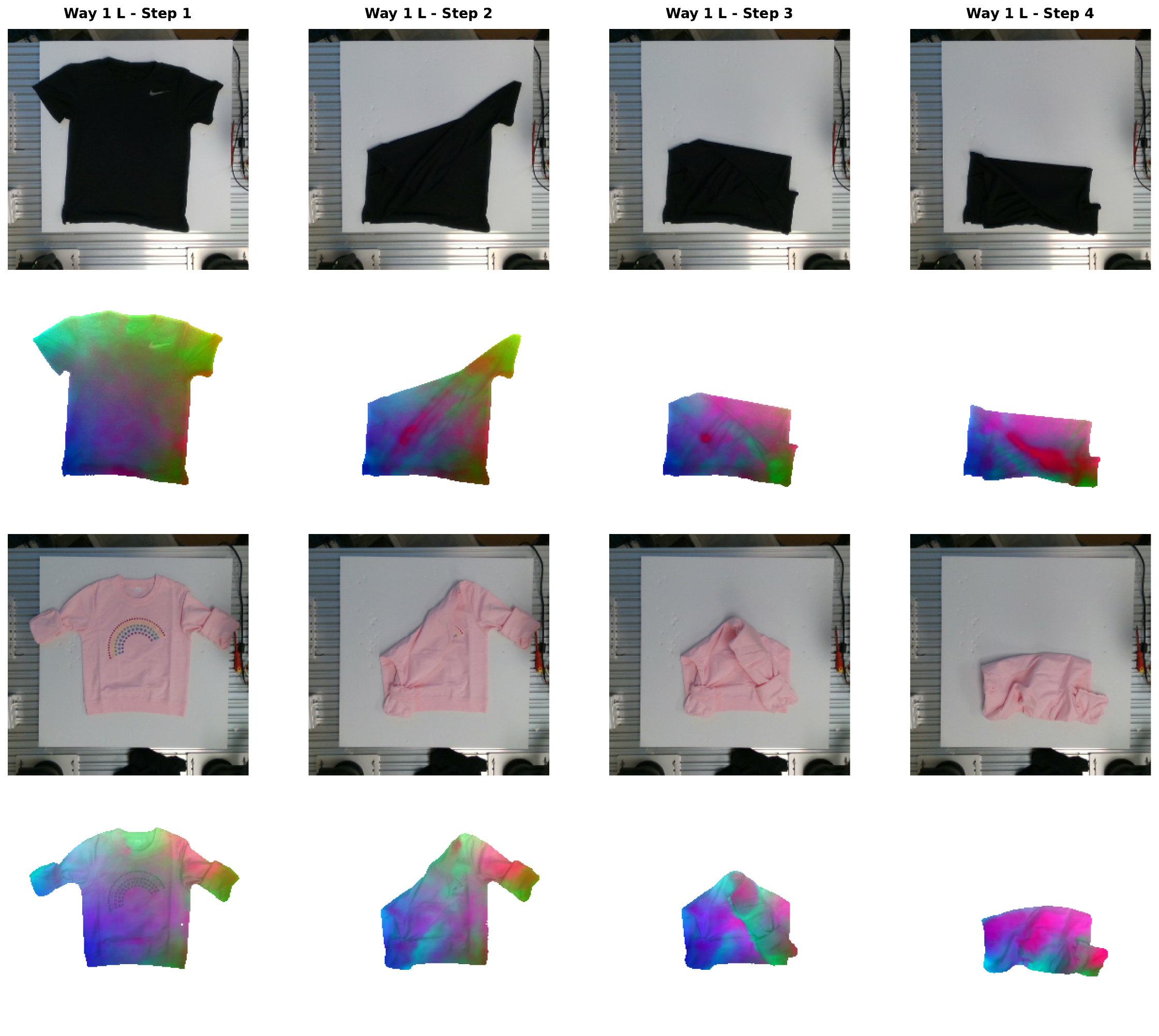}
    \end{minipage}\hfill
    \begin{minipage}[t]{0.49\textwidth}
        \centering
        \includegraphics[width=\linewidth]{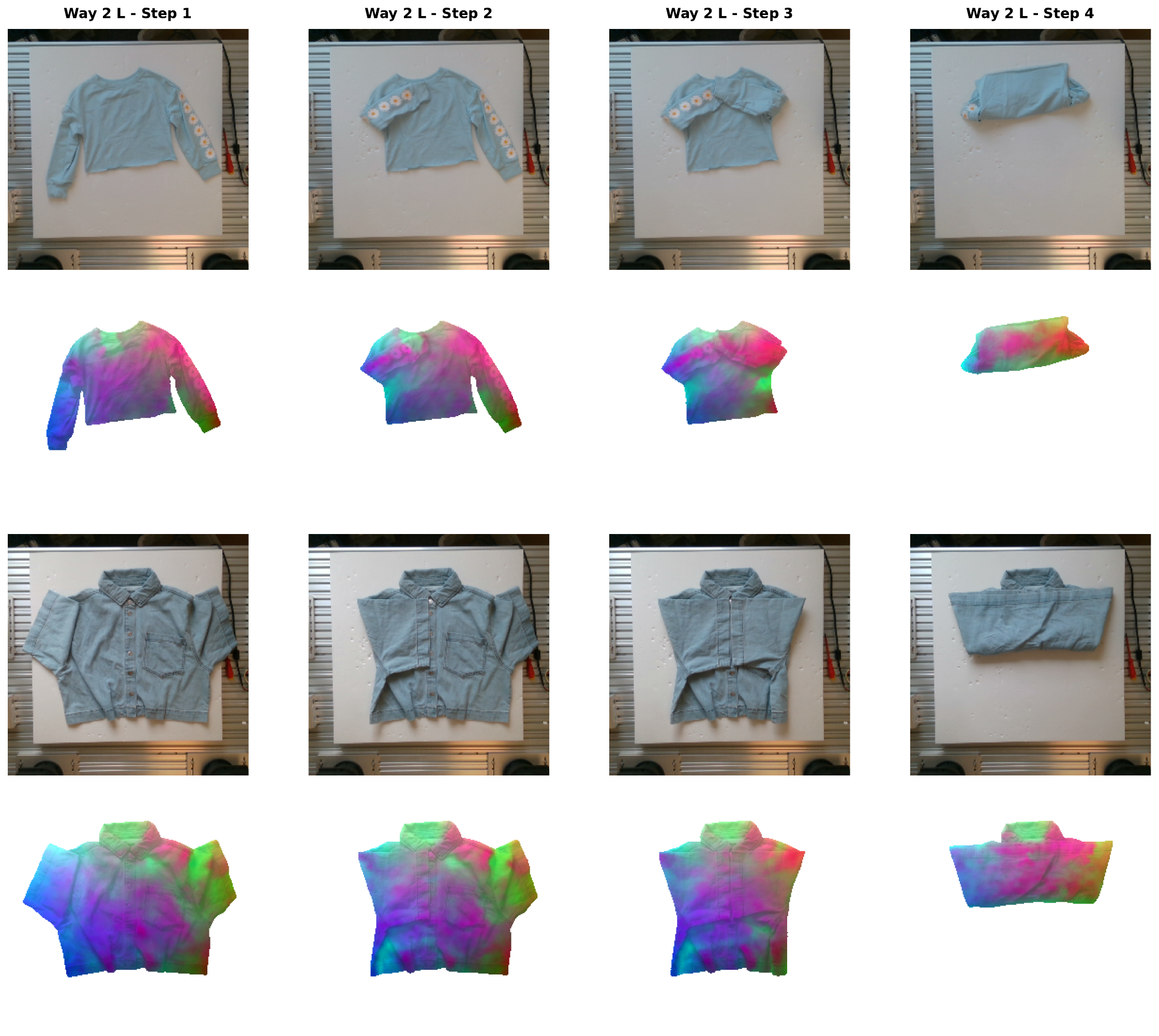}
    \end{minipage}

    \vspace{2mm}
    {\small (b) UniGarmentManip \cite{Wu_2024_CVPR} representations on real-world demonstrations.}

    \caption{Visualizations of DINOv3 and UniGarmentManip representations on real-world folding demonstrations. A shared PCA basis is fit to the pretrained token features and projected back onto the garment surfaces the same way as in Figure ~\ref{fig:realworld_pca_triptych}.}
    \label{fig:realworld_pca_triptych_compare}
\end{figure*}

\paragraph{Qualitative Comparison.}
Figures~\ref{fig:realworld_pca_triptych} and~\ref{fig:realworld_pca_triptych_compare} show visualizations of our encoder compared with raw DINOv3 (ViT-S/16+) \cite{simeoni2025dinov3} and UniGarmentManip \cite{Wu_2024_CVPR} representations on real-world folding demonstrations. Compared with raw DINOv3 features, our pretrained representation is markedly more temporally consistent: under the same PCA basis, DINOv3 often assigns different colors to the same garment regions across folding stages. This suggests that generic visual pretraining is less stable under large non-rigid cloth deformations than task-adapted representations. Compared with UniGarment's pretrained PointNet features, our representation is also smoother and more spatially coherent, whereas UniGarment features appear more mottled and locally noisy. Overall, our representation better preserves dense cloth geometry and semantic consistency across deformations.

\section{In-Context Policy Learning}
\label{architecture_details}

\begin{figure}[t]
    \centering
    \includegraphics[width=0.99\linewidth]{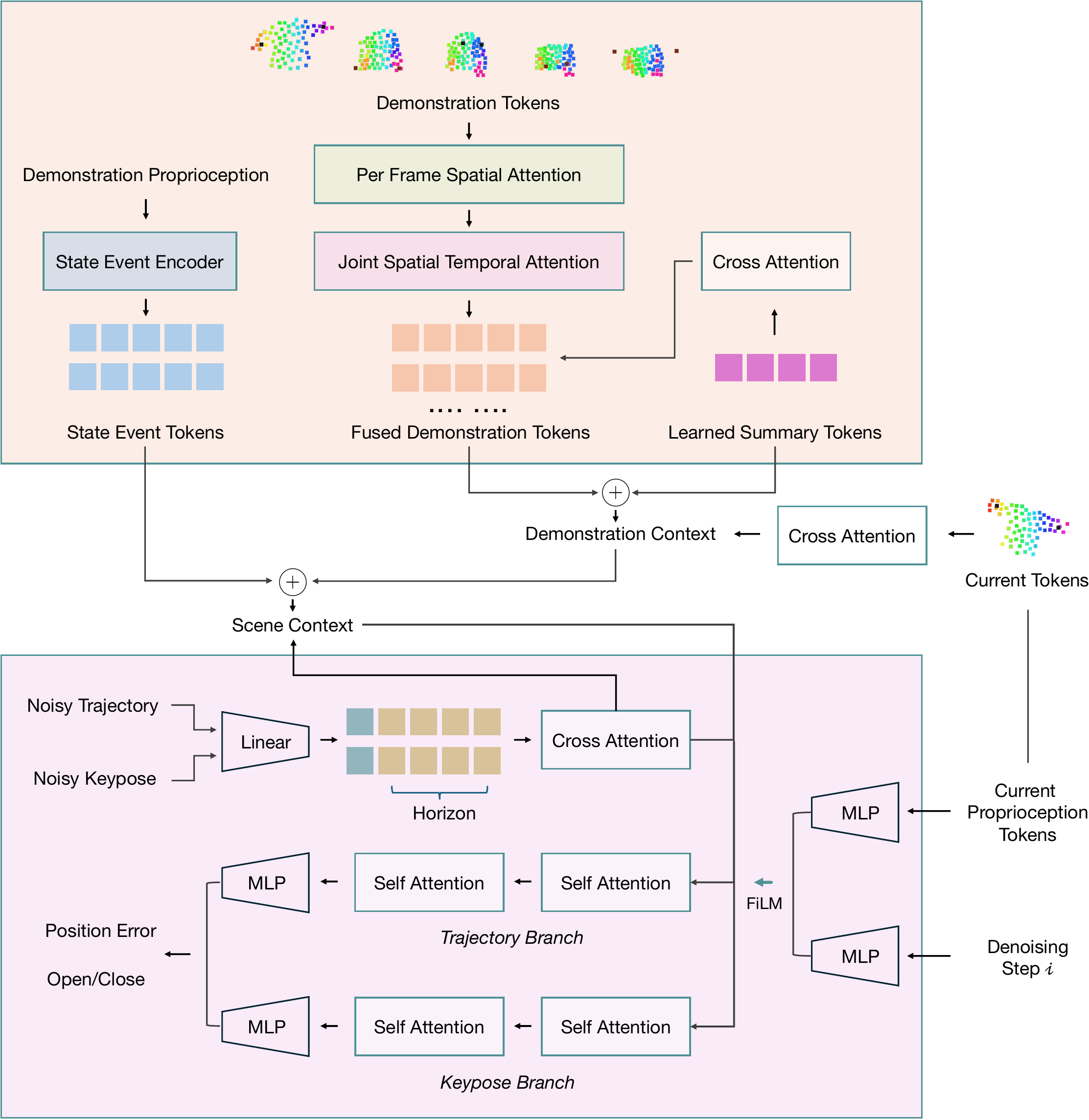}
    \caption{\textbf{Detailed architecture.} Top: the context encoder maps demonstration tokens into a demonstration context by combining cloth tokens and robot-state tokens in the context keyframes through per-frame spatial attention and joint spatio-temporal attention, followed by learned summary tokens. Bottom: the flow-matching action decoder fuses the current scene with the demonstration context, concatenates the resulting scene tokens with state-event tokens from a separate branch to form the scene context, and conditions noisy action tokens on this scene context, with denoising timestep and proprioceptive conditioning injected via AdaLN, to predict keypose and trajectory actions.}
    \label{fig:architecture}
    \vspace{-10pt}
\end{figure}

Figure~\ref{fig:architecture} details our in-context policy architecture, expanding on the overview in Figure~\ref{fig:overview}. It consists of a context encoder that transforms context keyframes into a demonstration context, and a flow-matching action decoder that predicts bimanual keypose and trajectory actions conditioned on the context-conditioned current observation, denoising timestep, and proprioceptive state.

\paragraph{Context Encoder Details.}
The context encoder processes both the demonstration keyframes and the current observation using shared visual and state tokenization modules. For each demonstration keyframe, a frozen pretrained cloth encoder lifts the RGB-D observation into camera-centric cloth tokens, which are projected into the policy embedding space. In parallel, each robot state is represented by a learned robot-state token constructed from the 4D end-effector state $(x,y,z,g)$, where $g$ denotes gripper openness. In our setting, each keyframe contains 64 cloth tokens and 2 robot-state tokens, one for each arm. Within each keyframe, these tokens are processed by 2 layers of per-frame spatial self-attention to model local cloth--robot interactions. Since both cloth and robot-state tokens are grounded in 3D, we equip these spatial-attention layers with RoPE and 3D ALiBi. RoPE encodes relative 3D geometry in the attention mechanism, while 3D ALiBi adds a distance-based inductive bias that encourages attention between spatially nearby cloth and robot tokens.

After per-frame spatial reasoning, the demonstration tokens are flattened across time and augmented once with temporal sinusoidal embeddings. The resulting sequence is processed by 4 layers of dense spatio-temporal self-attention, in which all tokens may attend to one another across both space and time. We adopt dense spatio-temporal attention rather than slot-wise temporal attention because the cloth tokens are obtained by FPS sampling in 3D, whose token ordering is not temporally consistent across frames, unlike fixed 2D patch grids. These dense spatio-temporal layers also use RoPE, but not 3D ALiBi. To summarize the temporally fused demonstration, we introduce 4 learned summary queries that cross-attend to the dense spatio-temporal demonstration tokens. The resulting summary tokens capture compact global mode information and are concatenated back to the token sequence, forming the final demonstration context that preserves both global intent and local geometry.

The current observation is encoded with the same frozen cloth encoder and robot-state tokenizer, producing current-scene tokens that consist of cloth tokens and current robot-state tokens. These tokens cross-attend to the demonstration context to become demo-conditioned scene tokens. In parallel, a separate state-event branch encodes the demonstration robot-state sequence with an MLP, adds hand-identity and temporal embeddings, and preserves gripper-state and end-effector motion information. The demo-conditioned scene tokens are concatenated with these state-event tokens to form the scene context. We find the state-event tokens to be important: without them, the model collapses all simultaneous folding modes such as mode1/S into sequential folds such as mode1/R.

\paragraph{Flow-Matching Action Decoder Details.}
The flow-matching action decoder builds on the base flow-matching transformer architecture of~\cite{gkanatsios20253d}; we additionally condition it on the scene context and introduce a keypose branch for short-horizon subgoal supervision, which we find critical for our long-horizon folding tasks. In the dual-arm setting, the noisy action input consists of two keypose tokens and $2H$ trajectory tokens for horizon $H$, where each token carries a 3D position. Keypose and trajectory tokens are first concatenated and embedded by a shared linear action tokenizer, then augmented with hand-identity and trajectory-time embeddings. These action-query tokens cross-attend to the scene context to produce context-conditioned action tokens, which are then split into keypose and trajectory branches. Each branch is concatenated with the scene context and refined by a self-attention block, followed by a second self-attention block for prediction refinement, before being decoded by four separate MLP heads predicting keypose position, keypose gripper openness, trajectory position, and trajectory gripper openness. The denoising timestep and proprioceptive state are encoded separately and injected via AdaLN: the timestep is embedded by an MLP, while the proprioceptive state is encoded by the shared state encoder, compressed into a single conditioning vector by a second MLP, and summed with the timestep embedding to form the AdaLN modulation signal. This conditioning is applied to the action-query cross-attention and all subsequent self-attention blocks. All attention blocks here use 3D relative attention, and keypose and trajectory tokens share the same denoising timestep during both training and inference.

\paragraph{Variable-Length Context Keyframes.}
Different folding modes may provide different numbers of context keyframes, so a batch may contain demonstrations of varying length. We pad each batch to the longest demonstration sequence and use a binary frame mask to indicate valid keyframes. This mask is expanded to token-level padding masks, ensuring attention ignores padded cloth,
robot state, and state-event tokens. During decoding, the masked demonstration tokens are concatenated with the current-scene and state-event tokens, and the same mask is passed to all attention blocks, allowing variable-length demonstrations to be processed in a single batch.

\section{Simulation Experiments}
\label{appendix_evaluation}

\paragraph{Training and Evaluation Details.}
For the results in Table~\ref{tab:policy_main_strict} and Table~\ref{tab:policy_ablation}, we train all models for 500 epochs (225k optimization steps) with the same random seed and report the last checkpoint for comparison. Training uses batch size 64. The policy uses a 384-dimensional transformer backbone, predicts a horizon of $H=4$, conditions on a single-step proprioceptive history, and performs 10 rectified-flow denoising steps. For relative action control, the dataset converts each trajectory into per-step position deltas from the current pose and normalizes them to $[-1,1]$ with a calibrated workspace min--max normalizer. The auxiliary keypose branch uses a separate normalizer, also calibrated on next-keyframe deltas, because keypose offsets are much larger than one-step motion. At test time, the policy is executed in a receding-horizon manner by rolling out 2 steps per query. We train with EMA and use the EMA weights for deployment and evaluation. The keypose auxiliary loss is weighted by 0.5. For each training sample, the dataset selects a random demonstration trajectory from the same mesh and folding mode as context conditioning. During inference, we encode the demonstration context once and cache it for reuse across subsequent policy queries. Each checkpoint is evaluated on 22 seen-context variants and 10 held-out-context variants across 60 held-out garments, for 1,920 rollouts in total. We fix the initialization and replay settings, including the same initialization protocol and sleeve-disturbance setup, to ensure fair comparison across methods. For oracle calibration, we generate 20 demonstrations per variant and garment under the same initialization protocol as the policy rollouts, yielding 38.4K oracle reference trajectories.

\paragraph{Evaluation Metric Details.}
All four metrics are computed in the \emph{same-init} setting. For each rollout, metrics are computed against the corresponding oracle trajectory, then aggregated per garment and reported as the mean with 95\% confidence intervals across garments.

\textbf{Ctx. Acc.} measures whether the rollout follows the requested folding context. We extract a 19D semantic-event descriptor from the rollout's first two grasp--release episodes and classify it with a source-matched 3-NN classifier over oracle descriptors from the same initialization family. A rollout is counted as correct if the predicted context matches the requested context, and Ctx.\ Acc.\ is the average of this binary outcome. \textbf{C-SR@95} measures joint semantic and physical success. A rollout is counted as successful only if it is first classified as the correct context and also passes the same-init oracle success envelope. This success envelope is calibrated separately for each source initialization and folding mode using oracle rollouts, and checks whether the rollout's final fold quality remains within the oracle 95th-percentile bound. In our implementation, this fold-quality test is composite: it depends on both final fold rate and final semantic geometry. \textbf{Geom.} measures final semantic geometric completion error. For each fold, we evaluate the mode-specific target keypoint errors at the final state, normalize each by the oracle median error for the same source initialization and folding mode, and take the worst normalized error. Thus Geom.\ asks how far the final folded garment is from the intended target geometry, with lower values indicating better completion. $\boldsymbol{W_1}$ measures distributional agreement with the oracle final geometry. For each context, we compute the Wasserstein-1 distance between the rollout and oracle distributions of the final semantic geometry coordinates, average across geometry dimensions, then average across contexts: \(W_1=\frac{1}{|\mathcal C|}\sum_{c\in\mathcal C} W_1(c)\). Unlike C-SR@95 and Geom., which evaluate each rollout individually, \(W_1\) evaluates whether the \emph{distribution} of outcomes produced by a method matches the oracle distribution.

\paragraph{Ablation Result Details.}
Table~\ref{tab:sameinit_ablation_raw} reports raw metrics for all ablations, showing that each major component improves held-out performance. The full model achieves the best held-out Ctx.\ Acc.\ (95.8), C-SR@95 (58.3), and $W_1$ (0.099), and near-best Geom.\ (1.89). Removing 3D ALiBi leaves seen performance almost unchanged (98.9 Ctx.\ Acc., 69.5 C-SR@95) but substantially hurts held-out generalization (84.0 Ctx.\ Acc., 43.5 C-SR@95, 2.19 Geom., 0.163 $W_1$), indicating that geometric bias is especially important for mode disambiguation under generalization. The largest held-out degradation comes from removing the keypose auxiliary (35.3 C-SR@95, 2.59 Geom., 0.182 $W_1$) or the context encoder (36.7 C-SR@95, 2.42 Geom., 0.156 $W_1$), showing that short-horizon subgoal supervision and explicit context encoding are both crucial for robust execution. Removing summary or state-event tokens also clearly hurts performance, reducing held-out C-SR@95 to 38.7 and 41.8, respectively, consistent with summary tokens encoding the global fold plan and state-event tokens preserving sparse grasp-release timing. Finally, removing proprioception from the context encoder mainly harms context-following and conditional success (75.7 Ctx.\ Acc., 43.7 C-SR@95) while leaving geometric quality largely intact (1.85 Geom., 0.105 $W_1$), suggesting that proprioceptive context matters more for choosing the correct fold than for refining final geometry.

\begin{table}[h]
\centering
\scriptsize
\caption{\textbf{Ablation raw metrics on 60 held-out garments over 32 folding contexts.}
Ctx. Acc. is context-following accuracy. C-SR@95 is conditional success: a rollout must follow the requested context and satisfy the oracle-calibrated 95th-percentile geometric success threshold. Geom. is semantic geometric completion error, and $W_1$ is the Wasserstein-1 distance to the oracle final state.}
\label{tab:sameinit_ablation_raw}

\resizebox{\columnwidth}{!}{
\begin{tabular}{lcccccccc}
\toprule
\textbf{Variant} &
\multicolumn{4}{c}{\textbf{Seen Folds}} &
\multicolumn{4}{c}{\textbf{Held-out Folds}} \\
\cmidrule(lr){2-5} \cmidrule(l){6-9}

& Ctx. Acc. $\uparrow$ & C-SR@95 $\uparrow$ & Geom. $\downarrow$ & $W_1$ $\downarrow$
& Ctx. Acc. $\uparrow$ & C-SR@95 $\uparrow$ & Geom. $\downarrow$ & $W_1$ $\downarrow$ \\
\midrule

Full
& \cellcolor{cvprsecond}98.9 $\pm$ 0.7
& \cellcolor{cvprbest}\textbf{69.8 $\pm$ 3.3}
& \cellcolor{cvprsecond}1.39 $\pm$ 0.08
& .041
& \cellcolor{cvprbest}\textbf{95.8 $\pm$ 2.0}
& \cellcolor{cvprbest}\textbf{58.3 $\pm$ 5.0}
& \cellcolor{cvprsecond}1.89 $\pm$ 0.69
& \cellcolor{cvprbest}\textbf{.099} \\
w/o context encoder
& 90.2 $\pm$ 0.6
& 62.3 $\pm$ 3.6
& 1.46 $\pm$ 0.11
& .045
& 74.7 $\pm$ 1.8
& 36.7 $\pm$ 3.8
& 2.42 $\pm$ 1.10
& .156 \\

w/o proprio in context
& 89.8 $\pm$ 0.7
& 59.5 $\pm$ 3.5
& 1.47 $\pm$ 0.10
& .045
& 75.7 $\pm$ 1.8
& \cellcolor{cvprsecond}43.7 $\pm$ 3.8
& \cellcolor{cvprbest}\textbf{1.85 $\pm$ 0.50}
& \cellcolor{cvprsecond}.105 \\

w/o keypose auxiliary
& 89.8 $\pm$ 0.7
& 63.9 $\pm$ 3.5
& 1.42 $\pm$ 0.09
& .044
& 73.8 $\pm$ 2.3
& 35.3 $\pm$ 4.2
& 2.59 $\pm$ 0.76
& .182 \\

w/o 3D ALiBI
& \cellcolor{cvprbest}\textbf{98.9 $\pm$ 0.6}
& \cellcolor{cvprsecond}69.5 $\pm$ 3.5
& \cellcolor{cvprbest}\textbf{1.38 $\pm$ 0.08}
& \cellcolor{cvprbest}\textbf{.038}
& \cellcolor{cvprsecond}84.0 $\pm$ 4.2
& 43.5 $\pm$ 5.1
& 2.19 $\pm$ 0.54
& .163 \\

w/o state-event tokens
& 90.5 $\pm$ 0.4
& 64.2 $\pm$ 3.3
& 1.42 $\pm$ 0.10
& .043
& 73.7 $\pm$ 2.3
& 41.8 $\pm$ 4.3
& 2.02 $\pm$ 0.64
& .125 \\

w/o summary tokens
& 90.8 $\pm$ 0.6
& 63.0 $\pm$ 3.3
& 1.44 $\pm$ 0.09
& \cellcolor{cvprsecond}.040
& 71.7 $\pm$ 2.3
& 38.7 $\pm$ 4.4
& 2.15 $\pm$ 0.64
& .133 \\

w/o context encoder \& keypose
& 88.9 $\pm$ 1.1
& 54.1 $\pm$ 3.8
& 1.65 $\pm$ 0.12
& .054
& 63.8 $\pm$ 2.4
& 28.7 $\pm$ 3.7
& 2.97 $\pm$ 1.49
& .201 \\

w/o state-event \& sum. tokens
& 89.8 $\pm$ 0.7
& 62.6 $\pm$ 3.4
& 1.42 $\pm$ 0.10
& .045
& 73.5 $\pm$ 2.2
& 41.0 $\pm$ 4.0
& 2.31 $\pm$ 1.25
& .111 \\

\bottomrule
\end{tabular}
}
\end{table}

\begin{figure}[h]
    \centering
    \includegraphics[width=0.995\linewidth]{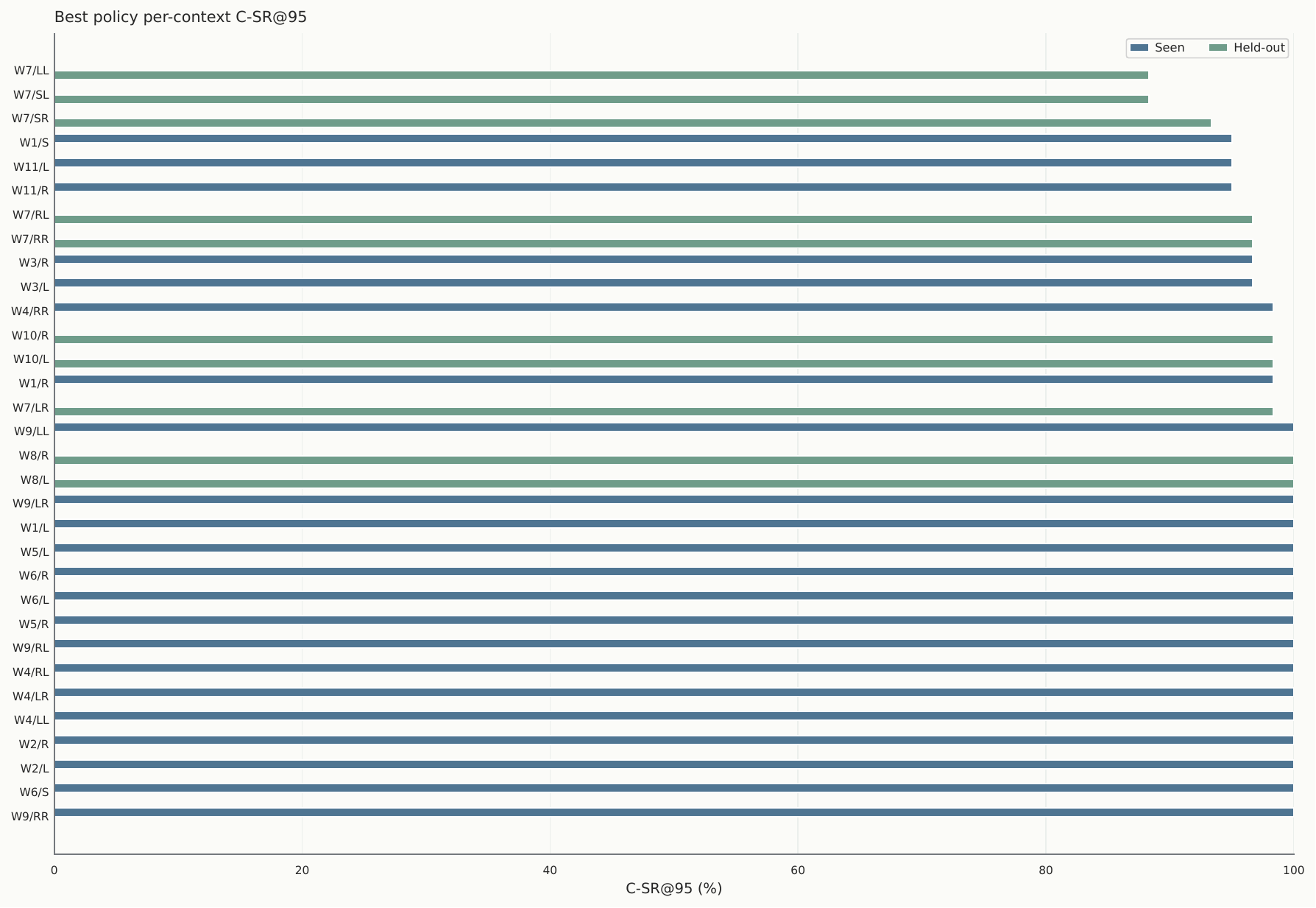}
    \caption{\textbf{Per-context same-init C-SR@95 for the best policy.} Bars are sorted by performance, with seen and held-out contexts shown in paired colors. The hardest held-out cases are the simultaneous/side-fold variants, especially \texttt{LL} and \texttt{SL}, followed by \texttt{SR}.}
    \label{fig:per-context-same-init}
\end{figure}

\paragraph{Per-mode Success Rates.}
Figure~\ref{fig:per-context-same-init} shows that the best policy is nearly saturated on seen contexts, with only a small drop on \texttt{S} compared with the other seen modes. On held-out contexts, the main bottlenecks are concentrated in the simultaneous and order-sensitive variants: \texttt{LL} and \texttt{SL} are lowest at 88.3, followed by \texttt{SR} at 93.3, while the remaining unseen modes are at or above 96.7. This suggests that the remaining failure cases are driven less by coarse task recognition than by fine two-arm coordination, especially when the policy must resolve grasp timing and placement order under cluttered sleeve geometry. In contrast, the non-simultaneous variants are almost fully solved.

\section{Real-world Experiments}
\label{Real-world-appendix}

\paragraph{Modelling Robot Occlusion.}
For real-world experiments, we observed severe overhead-camera occlusions, particularly during the second stage of folding. To mitigate this issue, we apply gripper occlusions to the current observation during both pretraining and policy learning, reducing the sim-to-real mismatch without regenerating the dataset. We use a position-dependent projection template to approximate robot-induced occlusions under the top-down camera view. The demonstration keyframes remain unoccluded. This encourages the model to rely on the visible cloth geometry, rather than occluded regions, when inferring actions.

\begin{figure}[h]
    \centering

    \begin{minipage}[b]{0.48\linewidth}
        \centering
        \includegraphics[width=\linewidth]{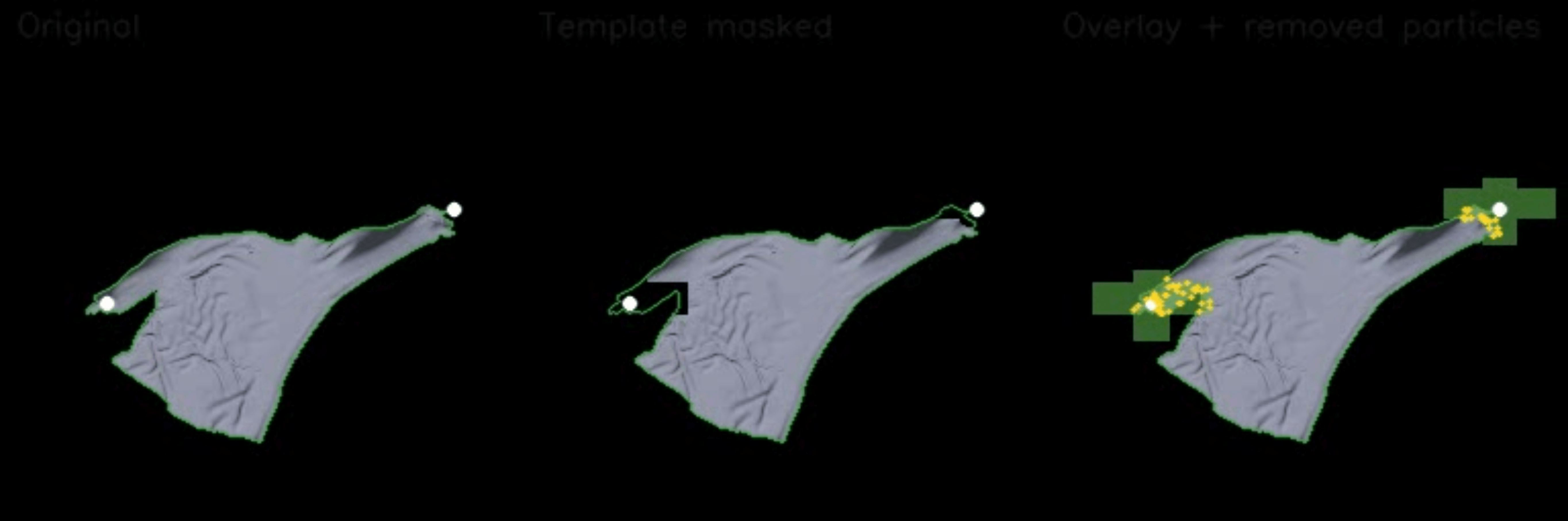}
    \end{minipage}
    \hfill
    \begin{minipage}[b]{0.48\linewidth}
        \centering
        \includegraphics[width=\linewidth]{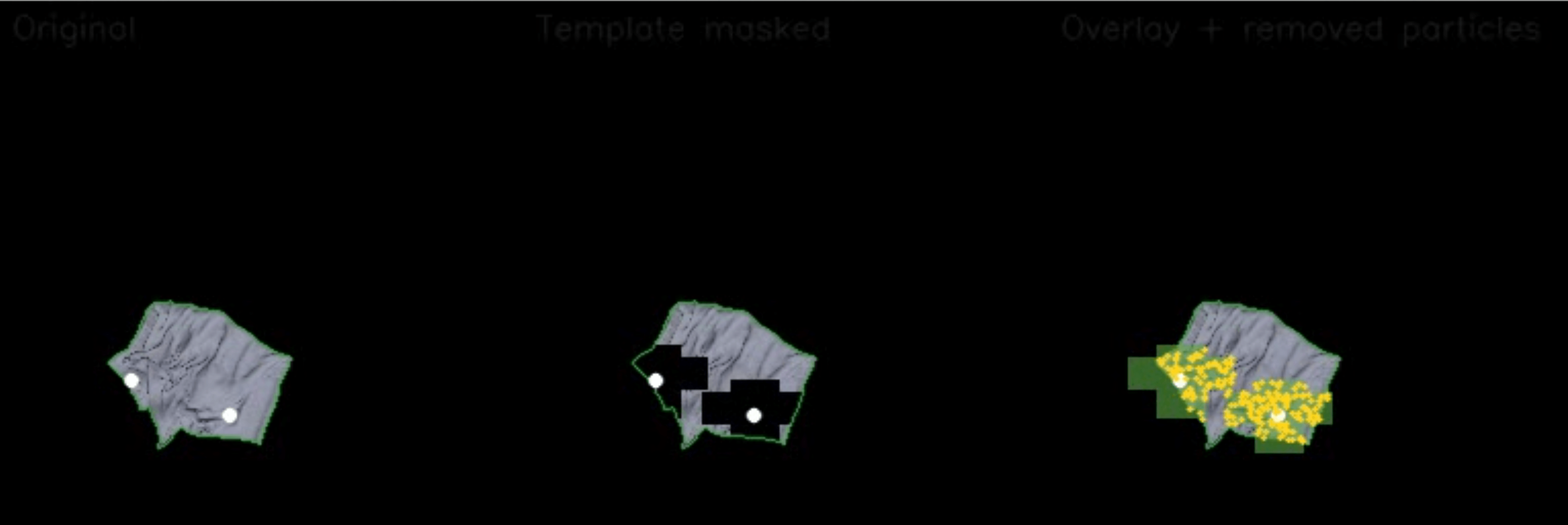}
    \end{minipage}
    \caption{\textbf{Gripper occlusion modelling.} Green contour: original cloth mask; white dots: gripper centers; yellow dots: newly occluded visible particles.}
    \label{fig:combined_occ}
\end{figure}

\paragraph{Training Details.} We use the best-performing model from simulation and train with an occlusion template radius of 3\,cm and a horizon of 8, as we found that longer horizons slightly improve robustness to robot-induced occlusions. To encourage recovery behaviors, we follow~\cite{vosylius2024instant} and add a 30\% probability of perturbing the gripper states, followed by relabeling recovery trajectories. We additionally randomize the initial gripper positions and open/close states before each task and, with 30\% probability, after first stage completion to encourage retry behaviors.

\paragraph{Policy Inference.}
During policy inference, we use a calibrated Intel RealSense D415 camera at a resolution of 640$\times$840 and adopt SAM2 \cite{ravi2025sam} for real-time cloth tracking using a bounding-box prompt. The segmentation model performs reliably in most cases, with occasional failures caused by robot-induced occlusions during the second stage of folding. To match the fixed aspect ratio used in simulation, we center-crop the images to a square resolution and adjust the camera intrinsics accordingly. As part of preprocessing, we filter out points that are either too close to or too far from the camera. The dual-arm end-effector states are transformed into the camera optical frame for policy inference and converted back to each robot's local frame for continuous execution using a Cartesian controller. The policy performs 10 denoising steps and executes the first 4 of 8 predicted actions at each iteration. We keep the end-effector orientations fixed throughout execution and linearly interpolate between actions to improve execution smoothness.

\paragraph{Cross-Embodiment Transfer.}
Since robot grippers are excluded from visual observations and all states and actions are expressed in the camera frame, the policy transfers across dual-arm embodiments without retraining. In Isaac Lab, we evaluate the policy on two Franka robots, while in the real world we deploy it on a Dobot platform. Demonstrations can be collected directly from human hands without robot teleoperation. We use the hand tracker from \cite{qian2025pianomime} to detect hand locations and project them onto preceding clean frames to recover gripper states at keyframes, followed by manual correction of missing or incorrect interaction points. For folding modes involving intermediate retraction (e.g., mode 2), the corresponding retraction points are obtained by interpolation. Each keyframe sequence is appended with a final folded state with both grippers at their home positions. In practice, only 3--4 demonstration keyframe images are required (Fig.~12), making demonstration speed and intermediate motions irrelevant. Hand tracking identifies interaction points correctly in about 90\% of cases; manual correction takes under 10\,s otherwise. Our soft contrastive representation makes the policy robust to small annotation errors.

\paragraph{Real-World Experiment Details.}
Figure~\ref{fig:real_world_clothes} shows the garments used in our real-world experiments, consisting of two shirts, two pairs of shorts, two blouses, one track jacket, and one denim jacket. Our training dataset is dominated by long-sleeve garments similar to \#1--2, contains a smaller proportion of garments resembling \#3--4, and includes few examples similar to \#5--8, making the latter garments rare or out-of-distribution. We observe that simple pick-and-place actions achieve limited success on garment \#8 due to its high stiffness, as the fabric does not naturally drape downward without additional smoothing or dynamic manipulation.

\clearpage

\begin{figure*}[h]
    \centering
    \includegraphics[width=0.905\textwidth,page=1]{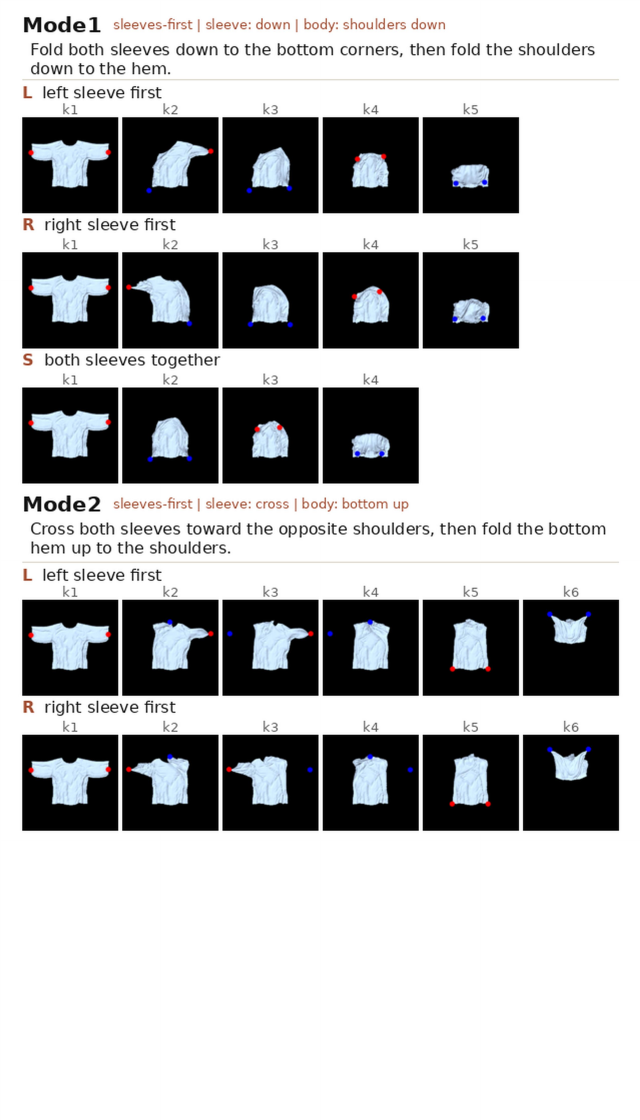}
    \caption{Visualization of all mode keyframes except for the final retraction frame.}
    \label{fig:strategy_bank}
\end{figure*}
\begin{figure*}[h]\ContinuedFloat
    \centering
    \includegraphics[width=0.905\textwidth,page=2]{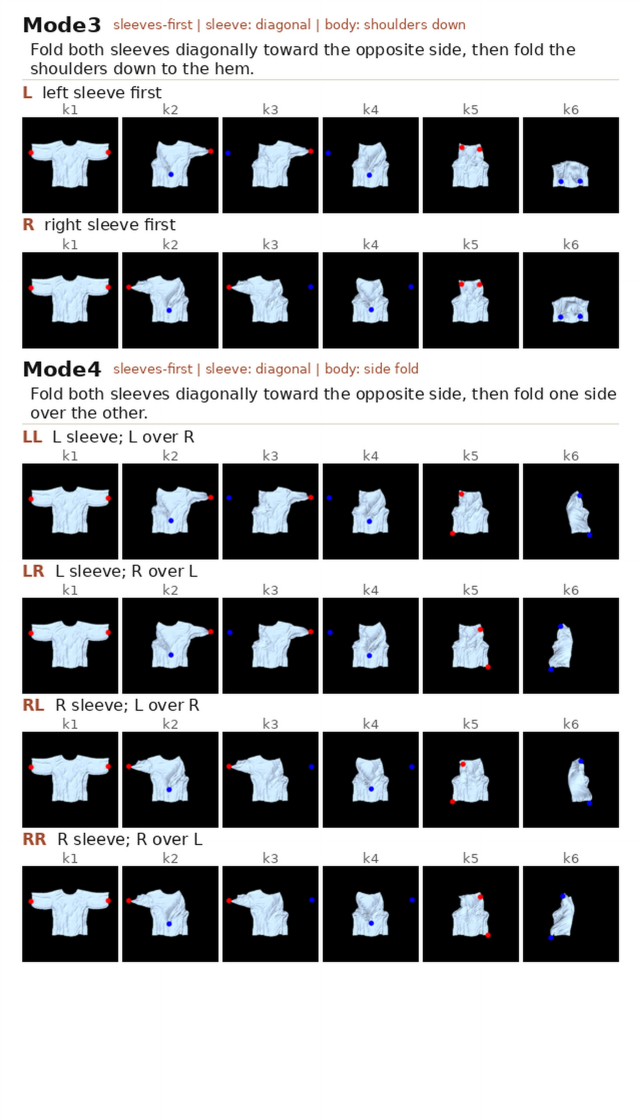}
    \caption{Visualization of all mode keyframes except for the final retraction frame (continued).}
\end{figure*}
\begin{figure*}[h]\ContinuedFloat
    \centering
    \includegraphics[width=0.905\textwidth,page=3]{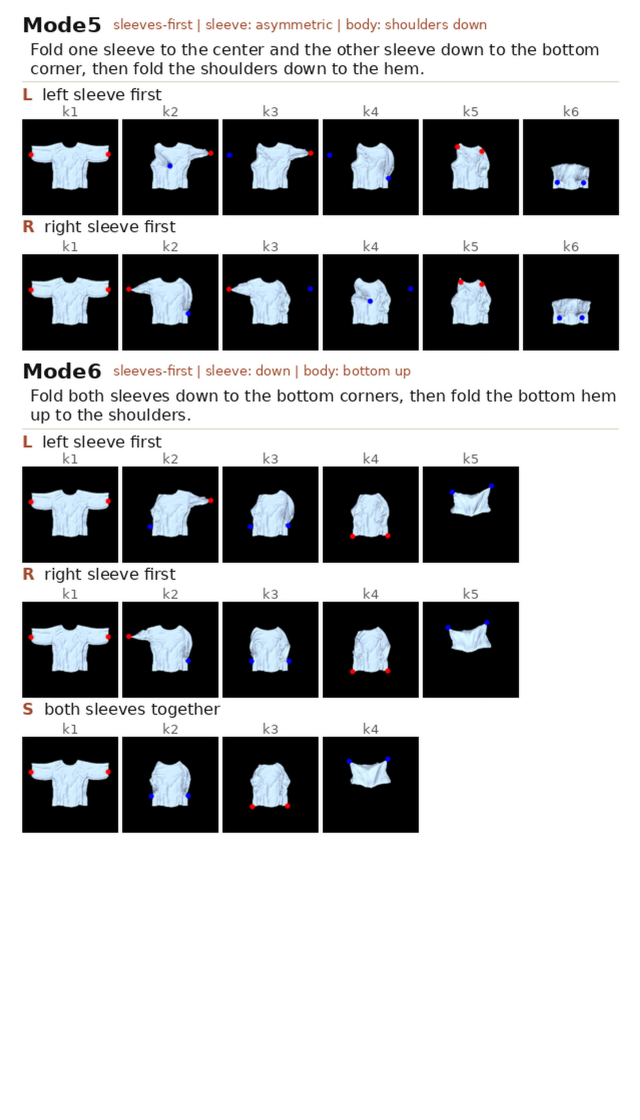}
    \caption{Visualization of all mode keyframes except for the final retraction frame (continued).}
\end{figure*}
\begin{figure*}[h]\ContinuedFloat
    \centering
    \includegraphics[width=0.905\textwidth,page=4]{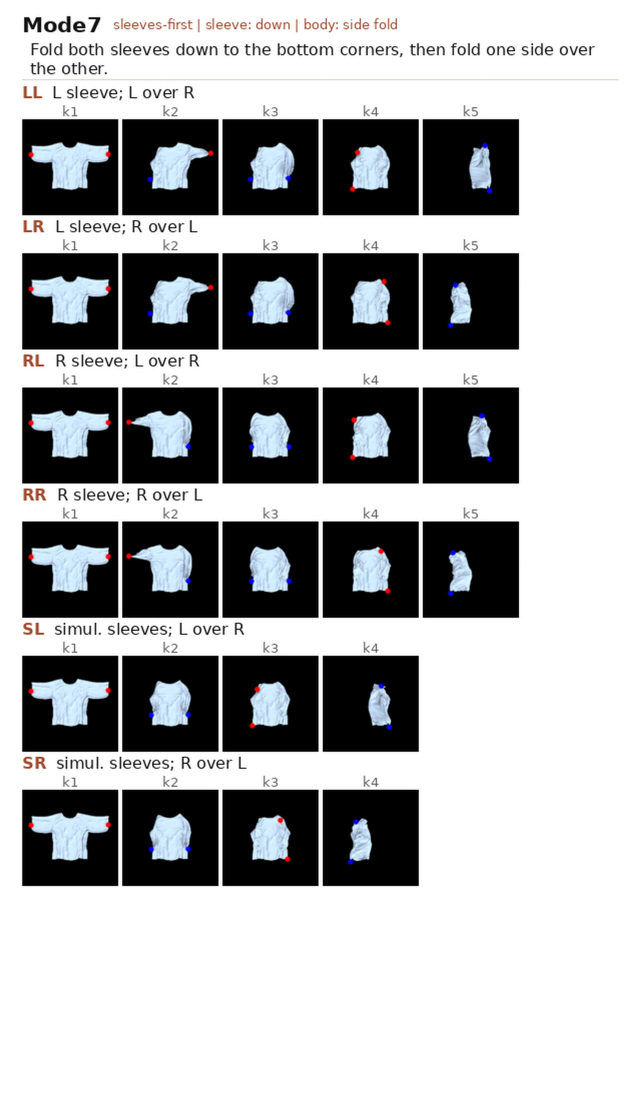}
    \caption{Visualization of all mode keyframes except for the final retraction frame (continued).}
\end{figure*}
\begin{figure*}[h]\ContinuedFloat
    \centering
    \includegraphics[width=0.905\textwidth,page=5]{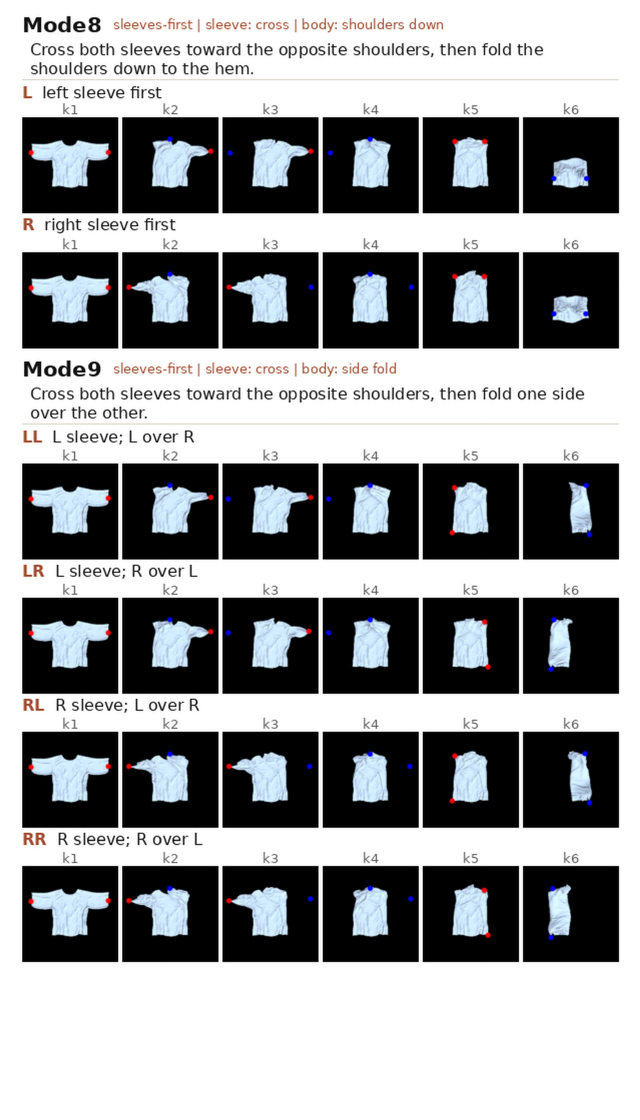}
    \caption{Visualization of all mode keyframes except for the final retraction frame (continued).}
\end{figure*}
\begin{figure*}[h]\ContinuedFloat
    \centering
    \includegraphics[width=0.905\textwidth,page=6]{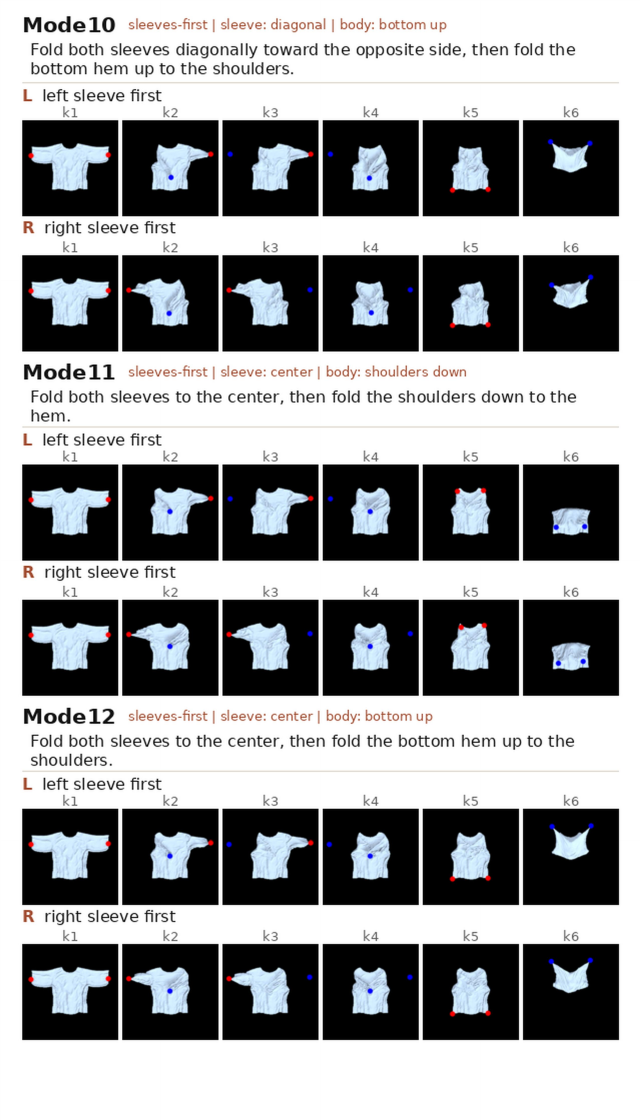}
    \caption{Visualization of all mode keyframes except for the final retraction frame (continued).}
\end{figure*}
\begin{figure*}[h]\ContinuedFloat
    \centering
    \includegraphics[width=0.905\textwidth,page=7]{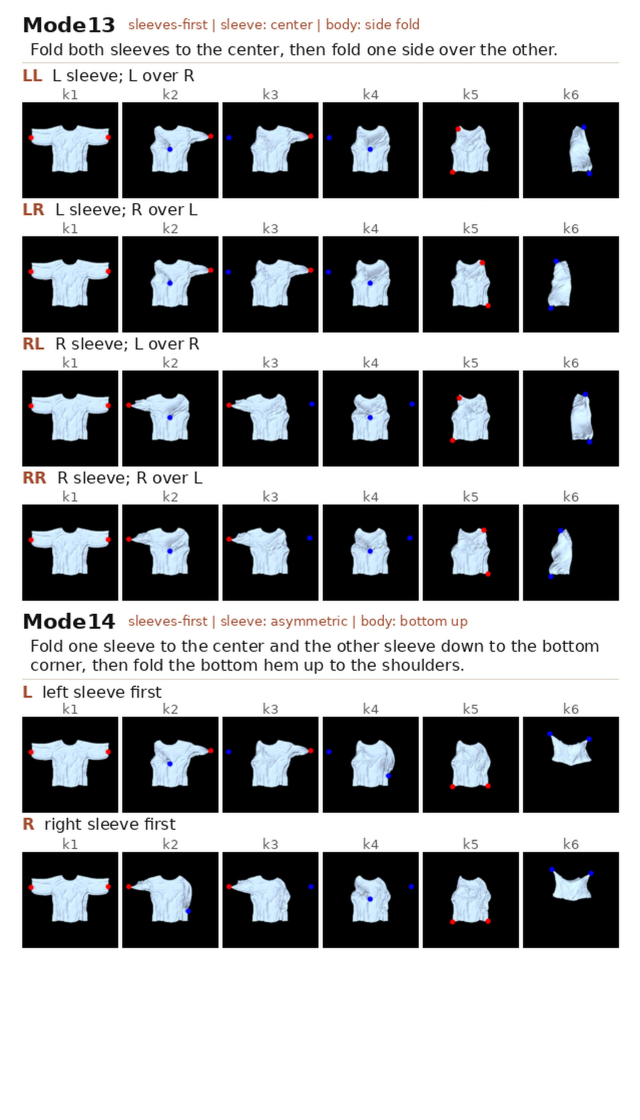}
    \caption{Visualization of all mode keyframes except for the final retraction frame (continued).}
\end{figure*}
\begin{figure*}[h]\ContinuedFloat
    \centering
    \includegraphics[width=0.905\textwidth,page=8]{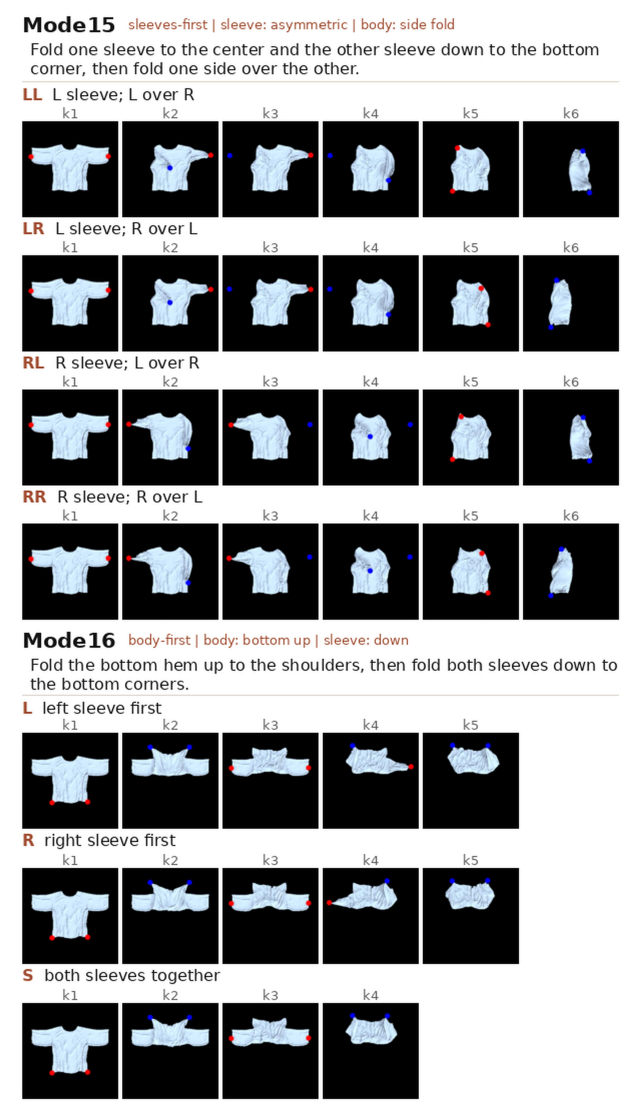}
    \caption{Visualization of all mode keyframes except for the final retraction frame (continued).}
\end{figure*}
\begin{figure*}[h]\ContinuedFloat
    \centering
    \includegraphics[width=0.905\textwidth,page=9]{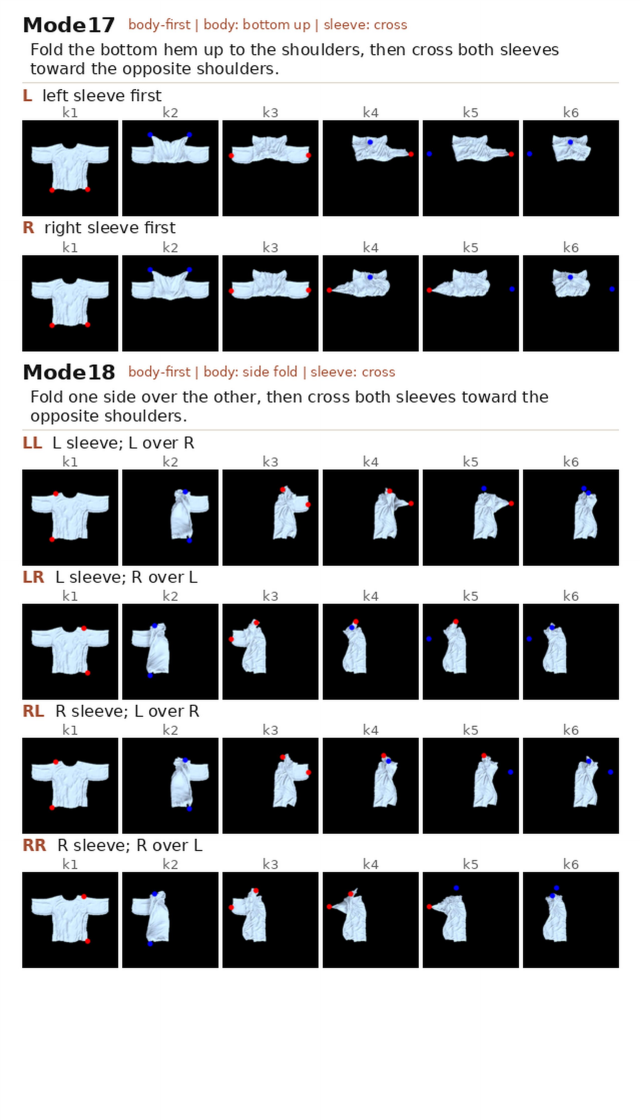}
    \caption{Visualization of all mode keyframes except for the final retraction frame (continued).}
\end{figure*}
\begin{figure*}[h]\ContinuedFloat
    \centering
    \includegraphics[width=0.905\textwidth,page=10]{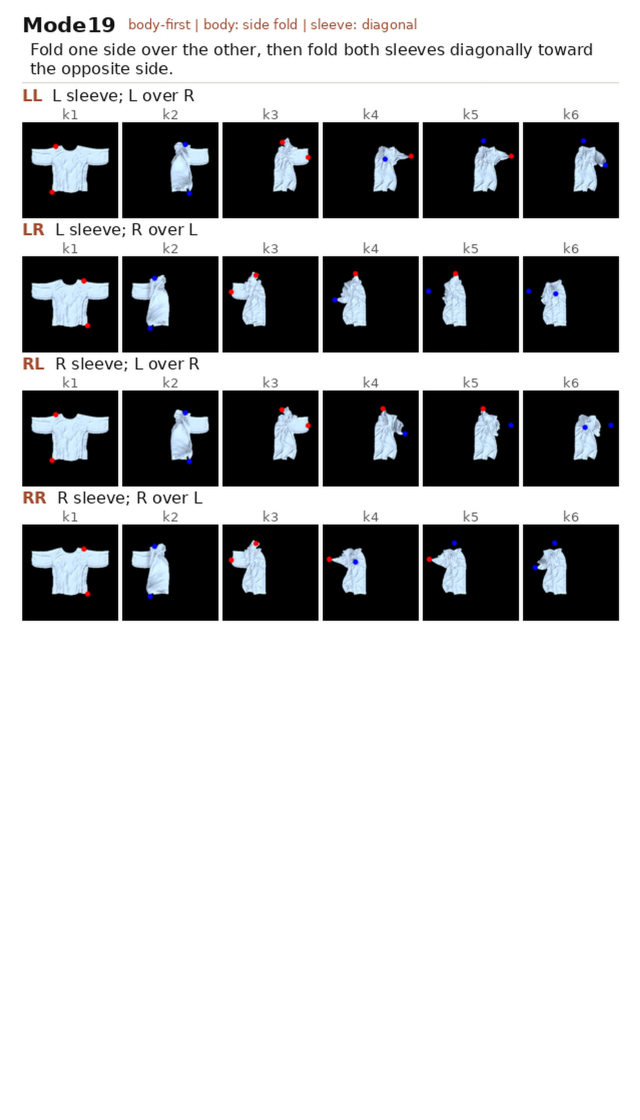}
    \caption{Visualization of all mode keyframes except for the final retraction frame (continued).}
\end{figure*}

\end{document}